\pdfoutput=1
\documentclass[sigconf,nonacm=true,screen=true,natbib=true]{acmart}

\setcopyright{none}
\renewcommand\footnotetextcopyrightpermission[1]{}
\usepackage{graphicx}
\DeclareGraphicsExtensions{.pdf,.png,.jpg}
\usepackage{pifont}
\usepackage{booktabs}
\usepackage{expl3}
\usepackage{tabularx}
\usepackage{array}
\usepackage{calc}
\usepackage{xparse}
\usepackage{textcomp}
\usepackage{textcomp}

\providecommand{\tightlist}{\setlength{\itemsep}{0pt}\setlength{\parskip}{0pt}}

\numberwithin{figure}{section}
\numberwithin{table}{section}

\begin{document}

\title{RecSys Factory: Bounding LLM Agent Autonomy to Decision Points in the Industrial Recommender Lifecycle}

\author{Dongyang Ao}
\authornote{Corresponding author.}
\affiliation{%
  \institution{FiT, Tencent}
  \city{Shenzhen}
  \country{China}}
\email{aodongyang@foxmail.com}

\author{Kaixiang Fang}
\affiliation{%
  \institution{FiT, Tencent}
  \city{Shenzhen}
  \country{China}}
\email{fkxworkmail@126.com}

\author{Shijie Xu}
\affiliation{%
  \institution{FiT, Tencent}
  \city{Shenzhen}
  \country{China}}
\email{xushijie09@gmail.com}

\begin{abstract}
Deploying LLM agents into industrial recommender operations
exposes a three-way tension we frame as the
autonomy--determinism--efficiency trilemma: general autonomy (interpreting operator
intent, generating glue code zero-shot), industrial determinism
(schema-conforming feature extraction, non-crashing A/B, zero
compliance-path hallucination), and end-to-end efficiency (an
author-recalled new-business-line onboarding compression on two lines
that we report as a case study rather than a generalization claim ---
see §6.3; controlled measurement pending v1.1). Any two can be maximized against the third; the failure modes at
each edge are documented in concurrent industrial-agent work. We present
\textbf{RecSys Factory}, an LLM-agent platform deployed for 78 days
across three heterogeneous Tencent recommender business lines. The
design principle is \emph{autonomy at decision points, not over
pipelines}, made concrete through three deconstructions that each
discharge one vertex of the trilemma. \textbf{Runtime} is deconstructed
into three host-emitted event sources (Claude Code \texttt{Stop} hooks,
the corporate-IM webhooks, the workflow scheduler APIs), giving
determinism: the platform carries no long-running daemon \emph{during
the wait phase} (a transient LangGraph process exists during the
$\approx$6\% of wall-clock spent on agent-side reasoning) and consumes
zero CPU during the 94\% of wall-clock spent waiting on Spark or GPU
jobs. \textbf{Capability} is deconstructed into a 29-file skill
ecosystem (8 971 lines of \texttt{SKILL.md}) whose per-skill pitfall
tables mechanically compile into a 400-entry \texttt{PitfallStore},
confining autonomy to bounded typed decision surfaces inside
pre-committed pipelines. \textbf{Deployment} is deconstructed across
three business lines with disjoint label semantics, A/B layer
topologies, and operator personas; the onboarding compression is
observed on two of the three (Business A recommendation and Business C
growth-marketing) as a case-study observation --- not a generalization
claim, and not measured against a controlled pre-platform baseline
(§6.3 flags this explicitly). The human is retained at
the diagnostic-vs-execution boundary via a corporate-IM
human-in-the-loop card protocol --- deployed as an audit-trail primitive
(schema-validated, idempotent, replayable) rather than a full HCI
contribution, and reported here from an 8-day 16-run pilot with
production rollout scheduled for v1.1. Across the
78-day window the platform recorded 1 624 CLI-tool dispatches (78.6\%
aggregate success rate counting \texttt{WAITING} outcomes as
non-success; $\approx$83.7\% if \texttt{WAITING} is treated as a correct
signal, i.e., end-to-end platform-error rate $\approx$16.3\%); a companion paper (AutoResearch, P3b)
instantiates the same substrate for autonomous research and produced
the IOSkip CIKM 2026 submission through a 196-round campaign on this
platform.
\end{abstract}


\maketitle

\hypertarget{introduction}{%
\section{Introduction}\label{introduction}}

A modern industrial recommender platform is not a single model but a
portfolio of pipelines that span sample generation, feature engineering,
model training, evaluation, A/B testing, and online serving. In a
typical Tencent business unit, three or more recommender business lines
(telecom payment recommendation, wealth-management product CVR,
credit-card scenario decisioning) run side by side, each with its own
labels, feature schema, A/B layer topology, and operator persona.
Algorithm engineers, business operators, and data analysts all need to
drive the same underlying machinery, but at different abstraction levels
--- engineers debug feature-store joins, operators tune item weights,
analysts attribute A/B effects across layers. The cognitive load is not
where most papers focus (model architecture); it is in the
\emph{operational substrate} --- knowing which weight to nudge, which
sample table to refresh, which causal-attribution layer to inspect when
a metric moves.

LLM agents promise to absorb this operational load. Deploying them into
industrial recommender operations, however, exposes a three-way tension
we frame as the
\textbf{autonomy--determinism--efficiency trilemma}.
\emph{General autonomy} --- the model's ability to interpret unstructured
operator intent, decompose a novel request into steps, and generate
glue code zero-shot --- is the reason to invoke an LLM at all; without
it a bash script suffices.
\emph{Industrial determinism} --- schema-conforming feature extraction,
non-crashing A/B rollouts, revenue-floor protection, and zero
hallucination on the compliance path --- is what production actually
requires; a 42\%-failure agentic pipeline
\cite{Vintschger2025} is not deployable regardless of how novel its
reasoning is.
\emph{End-to-end efficiency} --- compressing the new-business-line
baseline onboarding from a $\approx$14-day glue-code sprint to
$\approx$3 days of skill-authored setup on Business A and Business C
--- is what justifies the LLM overhead against a hand-crafted
alternative. We report this as an author-recalled case-study
observation, not a generalization claim: no controlled pre-platform
baseline was tracked, and §6.3 flags the same limitation for Business
C.

Any two vertices can be maximized against the third under a single
forcing resource: the \textbf{per-business-line engineering plus
human-oversight budget}, which at Tencent scale runs
$\approx$3--6 person-months to onboard a new recommender business line
and 40--80 review-hours per month to keep the resulting pipelines
running. A dollar spent hardening determinism (schema validation,
sandboxes, HITL gates, audit trails) is a dollar not spent authoring
skills (autonomy) or writing one-off glue code (efficiency); every
skill added to broaden autonomy is a skill that must be maintained,
regression-tested, and eventually reviewed by an engineer. The three
vertices therefore compete for the same budget rather than acting as
independent knobs, and each edge of the triangle names the failure
mode a system falls into when the budget forced it to abandon the
opposite vertex. Trading determinism
for autonomy plus efficiency is the failure mode of maximally agentic
systems: AI Scientist v2's free-form ReAct planner posts a 42\%
divergence rate on the MLE-Bench held-out set \cite{Vintschger2025}.
Trading autonomy for determinism plus efficiency is the failure mode of
classic workflow-fixed platforms like EasyRec \cite{Cheng2023AAAI} or
Monolith \cite{Liu2022Monolith} --- extremely stable and fast, but
schema drift or long-tail intent immediately re-engages a human
engineer. Trading efficiency for autonomy plus determinism is the
LLM-Copilot failure mode: Devin \cite{Cognition2024Devin}, OpenHands
\cite{Wang2024OpenHands}, or a review-every-line assistant will produce
correct code, but the multi-week paper-to-model tax is untouched
because every generated line still requires manual review before it
enters production.

\begin{figure}[t]
\centering
\includegraphics[width=\columnwidth]{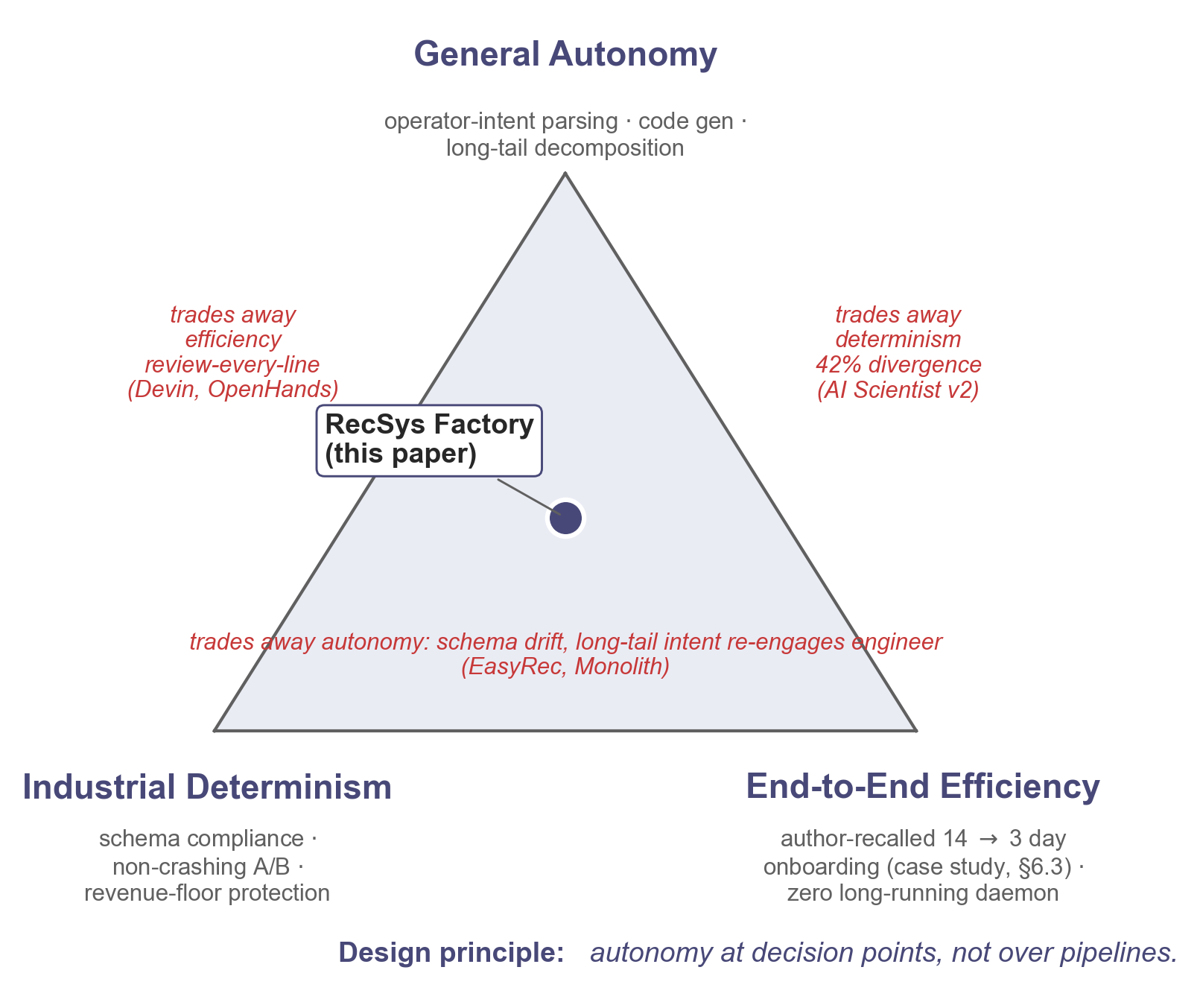}
\caption{The \emph{autonomy--determinism--efficiency trilemma} for
industrial LLM agents. Each vertex names a design goal; each edge names
the failure mode of pinning that edge (equivalently, of trading away
the opposite vertex). The three vertices compete under a single forcing
resource --- the per-business-line engineering plus human-oversight
budget ($\approx$3--6 person-months per new line at Tencent scale, plus
40--80 review-hours per month) --- which is what makes them pairwise
conflicting rather than three independent knobs. RecSys Factory commits
to an interior compromise
point rather than maximizing any single vertex, and instruments the
platform to make the compromise legible; the interior-compromise
principle collapses
the trilemma into three concrete engineering commitments described in
§3--§6.}
\label{fig:trilemma}
\end{figure}

Concurrent industrial-agent work responds to the same tension along
different axes. Kuaishou's AgentX \cite{AgentX2026} keeps the agent as
a self-evolving loop centred on one product surface; Tencent's NOVA
\cite{NOVA2026} attacks the determinism vertex with a
verification-aware harness but does not treat efficiency as an
independent variable. \textbf{RecSys Factory takes a different design
stance: instead of maximizing a single vertex, we commit to a specific
interior compromise point and instrument the platform to make the
compromise legible.} Figure~\ref{fig:trilemma} sketches the three
vertices, the failure mode of pinning any single edge, and our
compromise point.

\textbf{Design principle: autonomy at decision points, not over
pipelines.} The compromise becomes concrete through three
\emph{deconstructions} of the monolithic-agent stance, and each
deconstruction discharges one vertex of the trilemma.
\emph{Runtime is deconstructed} into host-emitted event sources
(enumerated below), which fixes the pipeline shape and gives us
\textbf{determinism}: the agent has no long-running process to
supervise \emph{during the wait phase} (only a transient LangGraph
process during $\approx$6\% agent-reasoning wall-clock) and inherits
the host's own uptime.
\emph{Capability is deconstructed} into a 29-file \texttt{SKILL.md}
ecosystem, each skill presenting the LLM with a small typed decision
surface (which sample table to refresh, which feature dimension to
compress, which override to propose) rather than a general reasoning
problem --- this is where we spend our \textbf{autonomy} budget, at
pre-committed decision points inside pre-committed pipelines.
\emph{Deployment is deconstructed} across three heterogeneous business
lines, stress-testing whether the compromise preserves
\textbf{efficiency} across recommendation, decision-support, and
growth-marketing operational regimes rather than just one flagship
product.

Around this compromise we adopt four quality-defence practices ---
schema validation at dispatch, sandboxed unit tests in the training
subgraph, a working A/B $\geq$ baseline$\cdot$80\% threshold, and
engineer-in-the-loop code review at the sensitive boundary --- as team
conventions the platform makes cheap enough to actually follow, not as
production hard gates. The human is retained at the
diagnostic-vs-execution boundary (§4's IM HITL card protocol): the
Analyzer proposes an override, the operator decides. Removing that
boundary erases the compromise; over-weighting it collapses efficiency.

\textbf{Contributions.} Each contribution defends one vertex of the
trilemma with a concrete engineering artifact and observable evidence.

\begin{enumerate}
\def\labelenumi{\arabic{enumi}.}
\item
  \textbf{Lifecycle-Aware Agent Framework (§3) --- defending
  determinism.} Three design principles turn the recommender lifecycle
  into first-class agent abstractions: (a) \emph{host-event coupling}
  --- the agent attaches to \texttt{Stop} hooks, the corporate-IM
  webhooks, and the workflow scheduler APIs, firing on host lifecycle
  events and sleeping otherwise; (b) \emph{single-source state} ---
  one \texttt{PipelineState} per run, persisted as inspectable JSON in
  SQLite, is the single truth across sessions; (c) \emph{skill
  subgraphs} --- domain knowledge is packaged as
  LangGraph-dispatchable subgraphs, not retrieval passages.
  Parasitic execution means \emph{no long-running daemon during the
  wait phase} (a transient LangGraph process runs during the $\approx$6\%
  of wall-clock devoted to agent-side reasoning; the remaining 94\% is
  actual zero-CPU wait). The platform ran 1 624 CLI-tool
  dispatches over 78 days without a scheduler process.
\item
  \textbf{29-Skill Ecosystem as Working Memory (§5) --- defending
  bounded autonomy.} 29 explicit skills totaling 8 971 lines of
  curated \texttt{SKILL.md} documentation, each a directory with a
  procedural body, bound script wrappers, and a machine-readable
  pitfall/problem/trap table. A rule-based extractor compiles the
  tables into a 400-entry \texttt{PitfallStore} (SQLite), which the
  platform's Plan node consumes at every planning step.
  the skill library becomes \emph{executable working
  memory} rather than a prompt library --- every operator gesture
  (``re-run business-A yesterday's training'') deterministically
  dispatches a single skill subgraph, and every documentation
  contribution (skill or changelog) mechanically enters the agent's
  runtime context at zero marginal effort. A business-specific skill
  onboarded post-launch (\texttt{domain-\allowbreak{}decision-\allowbreak{}c},
  business C) added 12 pitfall entries that the extractor absorbed
  without any code change.
\item
  \textbf{Three-Business-Line Deployment Case Studies (§6) ---
  reporting efficiency evidence in situ.} We report onboarding evidence
  from three heterogeneous business lines as case studies rather than
  as a generalization claim. Author-recalled onboarding time compressed
  from $\approx$14 days to $\approx$3 days on two of the three lines
  (recommendation product Business A and growth-marketing product
  Business C); no controlled pre-platform baseline was tracked, so this
  is a directional observation not a measurement.
  Business A (telecom personalization) reports +10--31 \% CPM lift
  across regional cohorts of one carrier plus +14--45 \% expected lift
  from rule-fallback identification on 9 sub-cohorts of a second
  carrier. Business B (telecom reranking decision support) reports a
  bootstrap 95 \% CI on the top-3 subset weight-tuning recommendation
  adopted by operators (P(\ensuremath{\Delta} \textgreater{} 0) = 100
  \%). Business C (wealth-management new-customer CVR audience-set
  orchestration) detected three simultaneous upstream data-integrity
  issues in a single pre-flight session that would have silently
  corrupted training. three heterogeneous business
  lines under one platform demonstrate that the trilemma compromise
  survives the operational realities of \emph{different} label
  semantics, A/B layer topologies, and operator personas.
\item
  \textbf{Silent-drift detection at production seams (§6.2, §6.3).}
  A methodological byproduct of embedding a ChatOps agent inside real
  business-line workflows is that the agent \emph{observes} points of
  silent divergence between what a configuration table says and what
  serving-time infrastructure does. In Business B we surface a
  ``configured weight $\ne$ effective weight'' diff (§6.2 C3) by
  reverse-engineering the effective weight from
  \texttt{(line\_score / raw\_pctr)} ratios in the daily
  snapshot --- exposing one item whose effective weight was 67$\times$
  lower than the operator-visible configured value. In Business C
  (§6.3) the pre-flight upstream-integrity check surfaces three
  simultaneous data-integrity failures in a single onboarding session.
  These are not features we designed for; they are what a platform
  earns when its agent is close enough to production seams to see
  them. Documenting the pattern is itself a contribution.
\end{enumerate}

\textbf{Human-in-the-loop as audit-trail primitive, not fallback.} The
corporate-IM card protocol described in §4 is not a graceful-degradation
channel: it is the explicit locus at which the Analyzer's diagnosis and
drafted override are recorded before an operator judgment is applied.
§4's protocol contributes an \emph{engineering} commitment
(schema-validated, idempotent, replayable) rather than an HCI evaluation;
the v1.0 statistics reported here are from an 8-day 16-run test-user
pilot (§4.4), and full production-operator HITL rollout is scheduled for
v1.1. §7 documents the two failure modes that
motivated retaining this boundary: (i) LLM diagnoses that were locally
coherent but violated business-line guardrails invisible to the agent,
and (ii) operator decisions whose rationale (upcoming campaign,
regulatory freeze) was never encoded in any skill or pitfall.

\textbf{Companion paper.} AutoResearch (P3b) instantiates the same
lifecycle-aware framework for autonomous research, contributing
surprise-weighted memory retrieval, two-tier knowledge sedimentation,
and a 5-mode memory ablation. The two papers share §3's framework
description and §2's related-work survey.

\hypertarget{related-work}{%
\section{Related Work}\label{related-work}}

We position RecSys Factory against four lines of work: industrial RecSys
frameworks, general-purpose LLM agent platforms, enterprise ChatOps and
multi-agent systems, and concurrent industrial RecSys agents. P3b
(AutoResearch) discusses the parallel literature on autonomous research
agents and agent memory architectures --- we cross-reference rather than
duplicate that section.

\hypertarget{industrial-recommender-frameworks-no-agent-layer}{%
\subsection{Industrial Recommender Frameworks (no agent
layer)}\label{industrial-recommender-frameworks-no-agent-layer}}

A first cluster establishes the \emph{modeling and infrastructure}
substrate that recommender platforms run on, but does not address
agent-level orchestration: EasyRec \cite{Cheng2023AAAI} (modular TF +
model zoo on Alibaba PAI), Monolith \cite{Liu2022Monolith}
(collisionless cuckoo-hash embeddings), Persia \cite{Lian2022Persia}
(hybrid sparse-async/sync-dense at 100T parameters), Angel
\cite{Jiang2018Angel} (parameter server), LiRank
\cite{LinkedIn2024LiRank} (unified ranking stack), and TransAct
V2/PinFM/PinRec \cite{Pinterest2025TransAct} (transformer ranking). All
report A/B-tested deployment but each sits \emph{under} the agent
layer --- they are the engines our agent dispatches against, not
competitors at the orchestration tier.

\hypertarget{llm-agent-platforms-general-purpose}{%
\subsection{LLM Agent Platforms
(general-purpose)}\label{llm-agent-platforms-general-purpose}}

A second cluster builds \emph{general-purpose} LLM agent infrastructure
without recommender-specific specialization: LangGraph and AutoGen
\cite{Wu2024AutoGen} provide state-machine and multi-agent
orchestration but ship without domain primitives; Devin
\cite{Cognition2024Devin}, OpenHands \cite{Wang2024OpenHands}, and
SWE-Agent \cite{Yang2024SWEAgent} target software engineering in
sandboxed environments --- precisely the architectural choice we depart
from when we couple to host-system events; MetaGPT
\cite{Hong2024MetaGPT} and ChatDev \cite{Qian2024ChatDev} coordinate
role-specialized agent teams for software-development tasks. None
target the recommender lifecycle or expose a chat interface for
non-engineer operators.

\hypertarget{enterprise-chatops-and-multi-agent-platforms}{%
\subsection{Enterprise ChatOps and Multi-Agent
Platforms}\label{enterprise-chatops-and-multi-agent-platforms}}

A third cluster builds general enterprise multi-agent platforms with
chat or DAG interfaces. JoyAgent-JDGenie \cite{JoyAgent2025} is the
closest enterprise comparator (open-sourced multi-agent ChatOps
reportedly serving 20K+ internal agents at JD), but treats
recommendation as a generic ``task'' with no domain primitives (no
CTR/CVR labels, no feature-store schemas, no A/B layer topology).
Academic ChatOps work --- ChatOps4Msa \cite{ChatOps4Msa2024}, IMPROVE
\cite{IMPROVE2025} --- proposes NL-fronted DAG execution but evaluates
on synthetic microservices, not recommender pipelines. RecSys Factory
is, to our knowledge, the first published agent platform that combines
ChatOps NL access, recommender-specific skill subgraphs, and
multi-business-line industrial deployment.

\hypertarget{llm-as-recommender-vs-llm-as-mlops-agent-disambiguation}{%
\subsection{LLM-as-Recommender vs LLM-as-MLOps-Agent
(disambiguation)}\label{llm-as-recommender-vs-llm-as-mlops-agent-disambiguation}}

A line sometimes confused with our setting uses LLMs as the reasoning
brain \emph{inside the recommender product itself}: InteRecAgent
\cite{Microsoft2023InteRecAgent} and RecMind \cite{Amazon2024RecMind}
expose LLM-based conversational recommenders to end users with
traditional recommenders as tools. Those systems live inside the
product surface (LLM = user-facing recommender). RecSys Factory
operates one layer above: LLM = engineering/operations agent that
drives the modeling pipeline. Complementary rather than comparable.

\hypertarget{concurrent-industrial-work}{%
\subsection{Concurrent Industrial
Work}\label{concurrent-industrial-work}}

Two closely-related industrial-agent papers arrived on arXiv shortly
before this submission. \textbf{AgentX} \cite{AgentX2026} (Kuaishou,
2026-06-26) is a 4-stage closed-loop production agent for Kuaishou App
feature engineering, with a Monitoring Platform for engineers and a
runtime playbook accumulated online. \textbf{NOVA} \cite{NOVA2026}
(Tencent, 2026-06-29) is a verification-aware harness for architecture
evolution on a billion-user ad ranking backbone, with L1--L4
task-complexity levels and an AutoRun/Copilot mode. Both target
production ranking systems at team-scale infrastructure with vertical
depth on one product surface. Table~\ref{tab:concurrent-comparison}
compares the three systems along five design axes to operationalize
the residual differentiation.

\begin{table*}[t]
\centering
\small
\begin{tabularx}{\textwidth}{@{}lXXX@{}}
\toprule
Axis & AgentX & NOVA & Ours \\
\midrule
\emph{Product surface} & feature engineering (candidate scoring) on one App feed & fine-ranking (pCTR/pCVR) on one ad backbone & three business lines: A recommender + B decision-support + C growth-marketing \\
\emph{Knowledge structure} & 4 evidence sources $\alpha$-weighted into prompt & static KB + trajectory memory $H$; failed candidates $\to$ forbidden directions & 29 dispatchable skill subgraphs + 400-entry PitfallStore mechanically extracted from 32 human-authored sources at deployment \\
\emph{Memory / retrieval signal} & 8-dim Quality Score threshold + adversarial-review gate for playbook admission & pattern-string hard-filter of failed candidates & surprise-DESC retrieval (residual $|$predicted \ensuremath{-} observed$|$) $+$ structural key \texttt{(baseline, modification\_class)} \\
\emph{HITL boundary} & Monitoring Platform (engineer-facing) & AutoRun for L1/L2, Copilot for L3/L4 & IM-card protocol at diagnostic-vs-execution boundary; approve/reject/custom with audit trail \\
\emph{Execution architecture} & Main-agent orchestrator (implicit daemon) & Main-agent orchestrator (implicit daemon) & Stop-hook lifecycle-coupled; zero long-running daemon during the wait phase (see §3.2, §3.5 caveats) \\
\bottomrule
\end{tabularx}
\caption{Five-axis comparison of concurrent industrial agent systems.}
\label{tab:concurrent-comparison}
\end{table*}

\textbf{When each design wins.} A team onboarding an additional
recommender business line whose failure mechanisms overlap existing
ones benefits most from RecSys Factory: the day-0 PitfallStore
short-circuits the ``blunder through first 50 experiments'' phase that
both concurrent systems rely on team curation to avoid. A team
maximizing offline AUC on a mature ad-ranking backbone benefits most
from NOVA's architecture-gradient formalism (modular search over
discrete architectural moves is more efficient than skill-dispatch
when depth on one system is the goal). A team pushing candidate-scoring
quality under a shared 24/7 monitoring surface benefits most from
AgentX's SGPO harness (closed-loop online playbook accumulation is more
expressive than our static PitfallStore when harness self-evolution is
primary). The three systems compose rather than compete; a systematic
head-to-head is v2 work.

\hypertarget{lifecycle-aware-agent-framework}{%
\section{Lifecycle-Aware Agent
Framework}\label{lifecycle-aware-agent-framework}}

This section describes the framework that grounds the \textbf{Skill
Ecosystem} (§5) and the \textbf{Three Business Lines deployment} (§6).
It is \emph{the} paper's main architectural contribution and is shared
in spirit --- though not in detail --- with our companion paper
AutoResearch (P3b §3, which describes the research-side instantiation of
the same framework).

\hypertarget{three-tier-architecture}{%
\subsection{Three-Tier Architecture}\label{three-tier-architecture}}

\begin{figure*}[t]
\centering
\includegraphics[width=\textwidth]{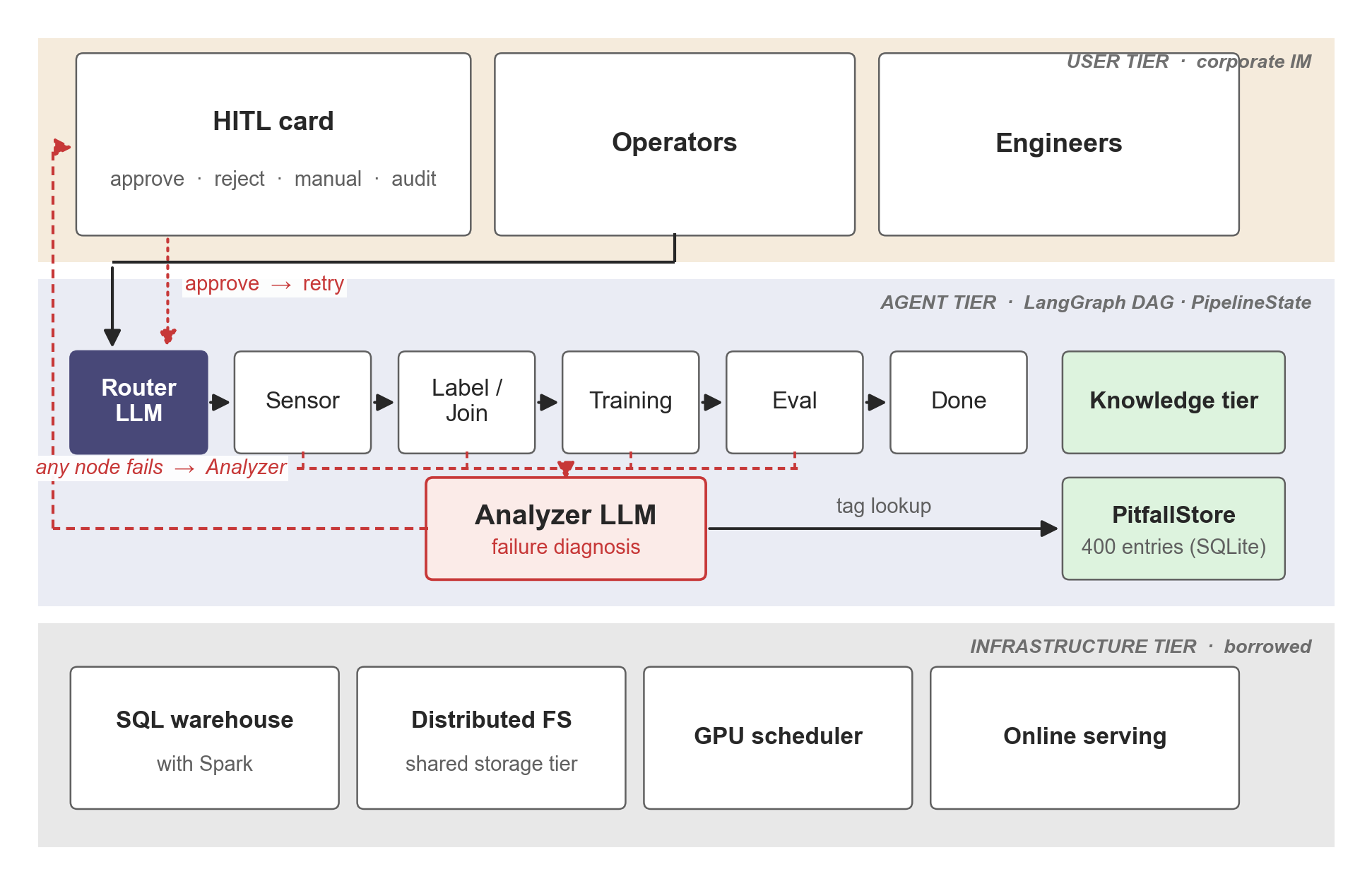}
\caption{\emph{Three-tier architecture with
lifecycle-coupled execution.} The User Tier surfaces Corporate-IM chat +
HITL card audit trail; the Agent Tier is a LangGraph DAG over an
immutable \texttt{PipelineState}, with a Knowledge Tier (SQLite) that
the Analyzer LLM queries by tag at diagnosis time; the Infrastructure Tier is
borrowed (distributed SQL / Spark / the shared storage tier / the GPU
scheduler / the online-serving platform). Dashed red arrows mark the
failure path (Train \ensuremath{\to} Analyzer \ensuremath{\to} HITL card
\ensuremath{\to} Router). Note. No tier owns a long-running process;
agent invocation is event-driven.}
\label{fig:arch}
\end{figure*}

RecSys Factory is decomposed into three loosely-coupled tiers that share
a single state object and communicate only through it (Figure~\ref{fig:arch}):

\begin{itemize}
\tightlist
\item
  \textbf{User Tier} --- Corporate-IM (the corporate IM) ChatOps
  surface. Operators (campaign managers, A/B reviewers) and engineers
  (algorithm, platform) submit natural-language requests through chat;
  the agent replies with structured human-in-the-loop (HITL) cards that
  carry an audit trail of which user approved which action at which
  time.
\item
  \textbf{Agent Tier} --- A LangGraph-based stateful DAG that decomposes
  a request into a sequence of skill invocations (sample construction,
  feature engineering, model training, evaluation, A/B attribution,
  decision support). Each node is a pure Python function over an
  immutable \texttt{PipelineState}.
\item
  \textbf{Infrastructure Tier} --- \emph{Borrowed}, not built: an
  internal GPU scheduler workflow engine for SQL / Spark / GPU
  scheduling, distributed SQL / Spark for sample tables, distributed FS
  / the shared storage tier for sample data, the online-serving platform
  for online serving. The agent's only writes are JSON-RPC
  \texttt{create\_task} / \texttt{start\_task} / the workflow API /
  \texttt{resource\ upload} calls; subsequent reads are pure polling.
\end{itemize}

The architecture's central design commitment is that \textbf{no tier
owns a long-running process}. The agent tier is invoked by lifecycle
events (IM webhook, HITL card callback, training-completion sentinel);
the infrastructure tier is the host platform's own scheduler; the user
tier is the end-user's own IM client. There is no agent daemon, no cron,
no message queue we own. This commitment --- ``lifecycle-coupled
execution'' --- is what lets the platform survive month-long deployment
windows without operator supervision.

\hypertarget{design-principle-i-host-event-coupling}{%
\subsection{Design Principle I --- Host-Event
Coupling}\label{design-principle-i-host-event-coupling}}

Conventional agent platforms (LangChain, AutoGen, MetaGPT) run the agent
loop inside a continuously-executing Python process. Sandboxed execution
platforms (Devin, OpenHands) run inside a long-lived Docker container.
Both choices require dedicated infrastructure (a process or a container)
for the agent itself. We instead \textbf{couple the agent to host-system
lifecycle events}:

\begin{itemize}
\tightlist
\item
  An IM webhook arriving at \texttt{POST\ /\allowbreak{}webhook/\allowbreak{}im} triggers a
  Router LLM call followed by a LangGraph DAG dispatch. When the DAG
  completes (success or HITL pause), the process exits.
\item
  A HITL card callback arriving at \texttt{POST\ /\allowbreak{}callback/\allowbreak{}im/\allowbreak{}card}
  resumes a paused DAG by reading \texttt{PipelineState} from SQLite,
  applying the operator's approval, and continuing from the paused node.
\item
  A training-completion sentinel file (written by the GPU scheduler at
  job completion) triggers the next post-training step via a Stop hook
  in the operator's CLI. (This sentinel mechanism is also AutoResearch's
  only out-of-band signal; see P3b §6.)
\end{itemize}

This pattern --- \textbf{parasitic coupling}, in the sense of consuming
the host's existing event grid rather than running our own --- is one of
the key reasons RecSys Factory deploys with operational footprint
comparable to a cron job rather than to a microservice.

\hypertarget{design-principle-ii-single-source-state-across-sessions}{%
\subsection{Design Principle II --- Single-Source State Across
Sessions}\label{design-principle-ii-single-source-state-across-sessions}}

A request that crosses a HITL gate or a long-running training job
\emph{must} survive process exit. We use a single Pydantic state object
(\texttt{PipelineState}) persisted as JSON in SQLite via
\texttt{aiosqlite}. The state carries: the request's lifecycle stage,
all task handles (the GPU scheduler \texttt{task\_flag},
\texttt{instance\_id}, \texttt{job\_id}, \texttt{exec\_id},
\texttt{yarn\_app\_id}), all \texttt{yaml\_\allowbreak{}overrides} accumulated
through HITL approvals, the \texttt{human\_\allowbreak{}approval} field that gates
retry-vs-end at the Analyzer node, and \texttt{dry\_run} /
\texttt{dry\_\allowbreak{}run\_\allowbreak{}scenario} flags that switch the executor to a mock
registry for offline testing.

Updates are immutable --- every node returns
\texttt{state.\allowbreak{}model\_\allowbreak{}copy(update=\{.\allowbreak{}.\allowbreak{}.\allowbreak{}\})} --- which means a
re-execution of any node from the persisted state is well-defined. This
in turn means the agent is \textbf{idempotent at the node granularity}:
a Webhook redelivery, a CLI restart, or a HITL callback retry never
corrupts in-flight work. The same property is what makes
\texttt{dry\_run} mode (used by \texttt{cli.py\ run}) trustworthy as a
regression test of the DAG itself.

\hypertarget{design-principle-iii-skill-subgraphs-as-composability-unit}{%
\subsection{Design Principle III --- Skill Subgraphs as Composability
Unit}\label{design-principle-iii-skill-subgraphs-as-composability-unit}}

Where AgentX organizes domain knowledge as four queryable evidence
sources blended via attention-style \ensuremath{\alpha}-weighting into
the prompt context, RecSys Factory organizes domain knowledge as 29
explicitly \textbf{dispatchable skill subgraphs}. A skill is a triple of
(a) a \texttt{SKILL.md} declaring inputs, outputs, side-effects, and
validation predicates; (b) a set of bound tool calls that implement the
skill (SQL templates, fid-xgb config templates, platform\_cli wrappers);
(c) optionally, sub-skills referenced by name for chained execution.

This distinction matters because operational recommender knowledge often
has a \emph{procedural} shape --- ``to attribute an A/B effect across
module-codes, run this 4-step procedure'' --- that is more naturally
expressed as a subgraph than as retrievable text. AgentX's
\ensuremath{\alpha}-weighted retrieval mixture is appropriate when
knowledge is fluid context; our dispatchable subgraph is appropriate
when knowledge is structured procedure. The two paradigms answer
different questions: AgentX asks \emph{what context to retrieve}, we ask
\emph{what subgraph to invoke}. We discuss the design's empirical trace
in §5 (skill-call statistics over the 10-week window, including hit
rate, average chain length, and cross-skill composition).

The subgraph formulation also gives us a research-autopilot \textbf{as a
special case}: the AutoResearch P2M subgraph (P3b §3) is itself a skill
registered in the same ecosystem, with \texttt{paper\_\allowbreak{}to\_\allowbreak{}model} as its
dispatch key. From the platform's perspective, autonomous research is
one more business line.

\hypertarget{recovery-semantics-and-adversarial-failure-modes}{%
\subsection{Recovery Semantics and Adversarial Failure
Modes}\label{recovery-semantics-and-adversarial-failure-modes}}

The framework's ``zero long-running daemon'' promise (§3.2) and
``single-source state'' promise (§3.3) invite a natural systems-review
question: what actually happens when the assumed invariants meet the
adversarial reality of distributed infrastructure? We enumerate the
concrete recovery-semantics contract v1.0 provides, and explicitly
name the failure modes for which no formal guarantee currently holds.

\textbf{Task submission (idempotency).} \texttt{create\_task}
receives a client-side deterministic \texttt{task\_flag} slug derived
from \texttt{(user, stage, run\_id, dispatch\_seq)}; on a redelivery
storm the workflow scheduler API rejects duplicate flags rather than
minting a second task, giving effective idempotency at the create
step. However the \texttt{start\_task} call that returns the
\texttt{instance\_id} is itself \emph{at-least-once}: a network retry
after a successful launch would in principle produce two identical
training runs. In v1.0 we detect this via a
\texttt{last\_instance\_id} field in \texttt{PipelineState} and abort
if a start-response arrives with a mismatching instance ID; a proper
exactly-once contract via server-side dedup is not in v1.0.

\textbf{State journaling.} \texttt{PipelineState} is persisted to
SQLite with \texttt{PRAGMA journal\_mode = WAL} and
\texttt{PRAGMA synchronous = NORMAL}; every state transition is
committed inside a single transaction. \texttt{state.json} snapshots
(used by AutoResearch's Stop-hook loop) are written to a temp file and
renamed via \texttt{os.replace} --- POSIX-atomic on the same
filesystem, but we do not enforce that both files live on the same
volume in v1.0. A torn write across mounts is possible; the loop
detects it by verifying a SHA-256 checksum on read and refusing to
resume with an unclear message rather than continuing with a partial
state.

\textbf{Callback and dual-account concurrency.} The corporate-IM card
callback endpoint carries a \texttt{card\_id} (payload-hash-deterministic)
and a \texttt{ballot\_state} (approve / reject / custom); the store
rejects a second callback for the same
\texttt{(card\_id, ballot\_state)} tuple, so retries are safe. Split
approvals across two channels are deduplicated at the endpoint but
not ordering-guaranteed --- the audit trail records both events; we
do not guarantee monotonic single-operator intent. Two concurrent
Stop hooks under the P3b §6.4 dual-account pattern serialize
optimistically via a compare-and-swap against
\texttt{(schema\_version, round\_counter)}; the loser retries after
re-reading, and a hook crashing between plan and write recovers on
the next fire. No cross-process lock; livelock-free empirically over
196 rounds but not formally proven.

\textbf{Where guarantees do not hold.} We explicitly do \emph{not}
claim: (i) exactly-once delivery of training-completion sentinels
across scheduler outages spanning $>$5 minutes (we depend on the
scheduler's own at-least-once semantics and treat duplicates as
idempotent no-ops when \texttt{state.json} shows the round already
committed); (ii) partition-tolerant serializability of the two-tier
memory reads (a rare cross-project write during a Plan-time retrieval
is served stale, then reconciled on the next round); (iii) formal
livelock-freedom of the dual-account CAS. Where these matter for
downstream users, we recommend running under a single-account
configuration until v1.1 tightens the invariants.

\textbf{Substrate mapping across the companion pair.}
RecSys Factory's ChatOps DAG (§3.6 below) and AutoResearch's P2M
subgraph (P3b §3) share this framework but consume different
combinations of the host-emitted event sources.
Table~\ref{tab:substrate-mapping} enumerates the assignment; the
operator-triggered ChatOps DAG is a ``poll-then-diagnose'' pattern
because it is invoked by an external IM message and then polls the
scheduler at each stage, while the autonomous P2M loop is a
``Stop-hook parasitism'' pattern because it self-triggers via the
Claude Code Stop hook and polls the scheduler only inside a single
round. The two patterns share the same event grid; they differ in
which vertex of the grid supplies the round-boundary trigger.

\begin{table*}[t]
\centering
\small
\begin{tabularx}{\linewidth}{@{}X l X X@{}}
\toprule
Event source & DAG & Consuming node & Semantic \\
\midrule
IM webhook \texttt{POST /webhook/im} & ChatOps & Router $\to$ Sensor & trigger \\
HITL card callback \texttt{POST /callback/im/card} & ChatOps & Analyzer / Train\_GPU & resume \\
Training-completion sentinel (GPU scheduler) & ChatOps & post-training step & trigger (poll-then-diagnose) \\
Training-completion sentinel (GPU scheduler) & P2M & WaitResult $\to$ Reflect & poll (Stop-hook parasitism) \\
Claude Code \texttt{Stop} hook & P2M & Reflect $\to$ Plan $\to$ Submit & trigger (self-triggering research loop) \\
Workflow-scheduler status poll & Both & Sensor (ChatOps) / WaitResult (P2M) & poll \\
\bottomrule
\end{tabularx}
\caption{Substrate mapping for the two DAGs served by the same
framework. The ChatOps DAG (P3a §3.6) is externally triggered by
operator IM messages; the P2M subgraph (P3b §3) is
self-triggered by \texttt{Stop} hooks. Both share the workflow-scheduler
status poll but differ in what supplies the round-boundary trigger.
\label{tab:substrate-mapping}}
\end{table*}

\hypertarget{the-chatops-dag}{%
\subsection{The ChatOps DAG}\label{the-chatops-dag}}

The agent tier's main DAG realizes the standard recommender
pipeline with HITL fallback (see Figure~\ref{fig:arch} for the tier
topology). Sensor polls the GPU scheduler for upstream job completion.
Sample-Labeling and Feature-Join dispatch SQL templates to a
general-purpose Spark workflow via the workflow API. Training and
Evaluation submit JSON-RPC tasks via the workflow scheduler API. On any
node failure, control transfers to \texttt{Analyzer} --- the LLM-driven
diagnosis node --- which fetches a two-layer log (the scheduler-log
service for scheduler diagnostics, YARN for container diagnostics) and
emits a structured \texttt{diagnosis} plus a candidate
\texttt{suggested\_\allowbreak{}overrides} payload. The diagnosis is rendered as a
HITL card; the operator's \texttt{approve} decision merges the overrides
into \texttt{yaml\_\allowbreak{}overrides} and re-routes the DAG back to Training,
while \texttt{reject} ends the run with a recorded failure record.

We detail the HITL contract --- request format, card schema, callback
semantics, and the 10-week + 8-day-pilot statistics on operator approval
rate and Analyzer diagnostic accuracy --- in §4.

\hypertarget{relation-to-existing-agent-platforms}{%
\subsection{Relation to Existing Agent
Platforms}\label{relation-to-existing-agent-platforms}}

Compared with \textbf{LangGraph} alone (which is the orchestrator we
use), RecSys Factory adds a domain-specific node library, a HITL card
protocol, persistent state across sessions, and the host-event coupling
pattern of §3.2. Compared with \textbf{AutoGen} \cite{Wu2024AutoGen} and
\textbf{MetaGPT} \cite{Hong2024MetaGPT}, we share the multi-step
orchestration pattern but discard the role-specialized agent metaphor
--- our nodes are pipeline stages, not personas. Compared with
\textbf{Devin} \cite{Cognition2024Devin} / \textbf{OpenHands}
\cite{Wang2024OpenHands}, we discard the sandboxed-Docker abstraction in
favor of explicit infrastructure-tier coupling --- the recommender
lifecycle \emph{is} the substrate we couple to, not something to be
reproduced inside a sandbox.

\hypertarget{chatops-dag-with-human-in-the-loop-cards}{%
\section{ChatOps DAG with Human-in-the-Loop
Cards}\label{chatops-dag-with-human-in-the-loop-cards}}

§3.5 sketched the LangGraph DAG that orchestrates a training request;
this section specifies the \textbf{request contract}, the \textbf{HITL
card protocol}, the \textbf{Analyzer diagnosis node}, and the 10-week
operational statistics plus the 8-day HITL DAG pilot statistics that
quantify each of these. We consider §4 the \emph{operator-facing} face
of the framework: everything a non-engineer campaign manager or A/B
reviewer perceives lives in this section, which is why we treat it as
first-class rather than as an appendix to §3.

\begin{figure*}[t]
\centering
\includegraphics[width=\textwidth]{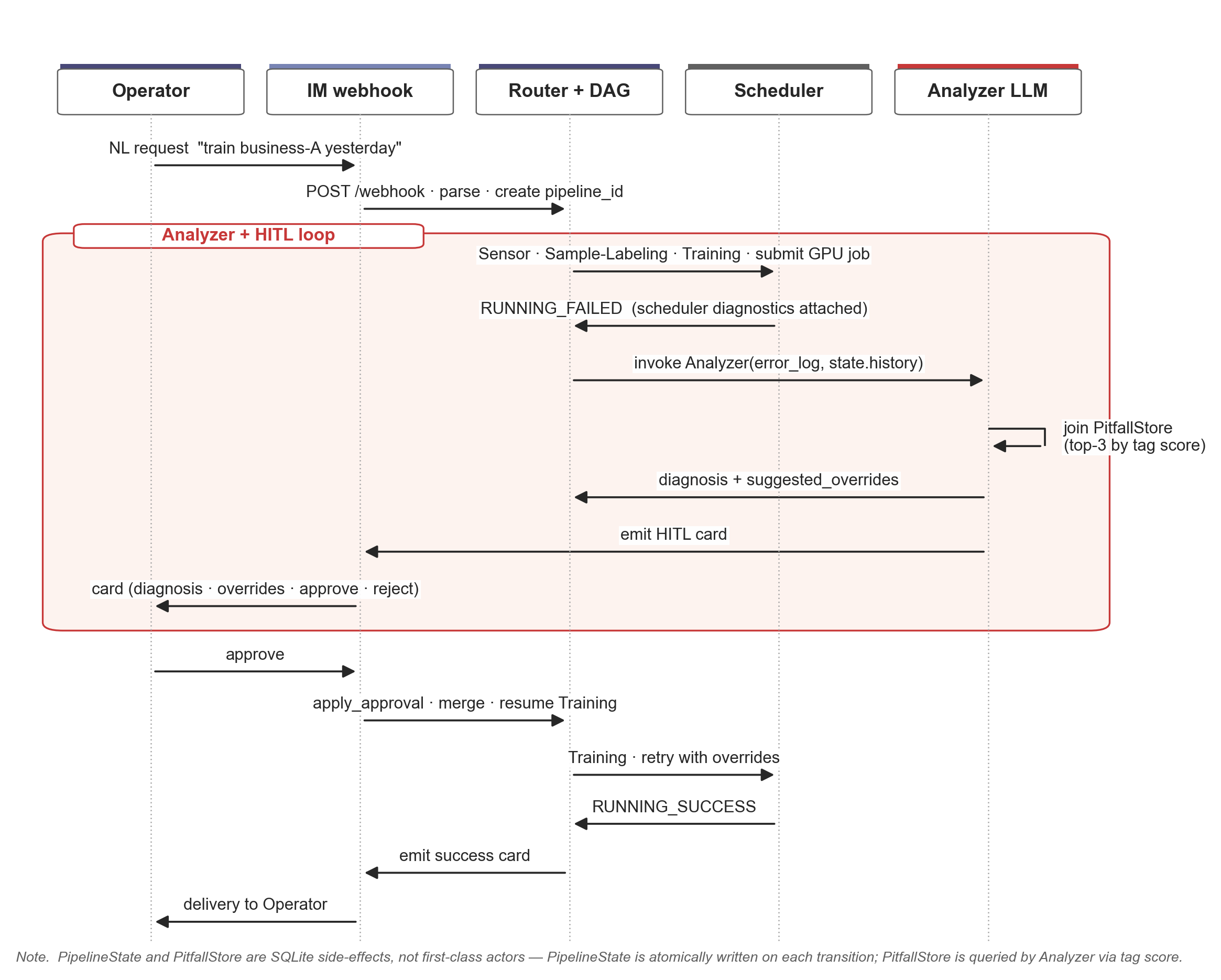}
\caption{\emph{End-to-end the corporate IM-to-HITL-card
lifecycle.} An operator's natural-language request enters through the IM
webhook, the Router LLM dispatches a LangGraph DAG that submits the GPU
scheduler tasks, and on failure the Analyzer LLM produces a structured
diagnosis fetched against \texttt{PitfallStore} before emitting a HITL
card. The operator's \texttt{approve} merges suggested overrides into
\texttt{PipelineState} and resumes the DAG at Training. The highlighted
band (steps 10--17) is the Analyzer + HITL loop that separates diagnosis
from execution. Note. Numbering is autonumbered by sequence order, not
by DAG topology.}
\end{figure*}

\hypertarget{request-contract}{%
\subsection{Request Contract}\label{request-contract}}

Any IM message routed to the platform is intent-classified by a
\textbf{Router LLM call} into one of six top-level intents:
\texttt{train\_\allowbreak{}new\_\allowbreak{}model},
\texttt{retry\_\allowbreak{}failed\_\allowbreak{}job},
\texttt{explain\_\allowbreak{}result},
\texttt{attribute\_\allowbreak{}ab\_\allowbreak{}effect},
\texttt{dispatch\_\allowbreak{}skill},
and \texttt{unknown}. The Router then emits a JSON envelope
(\texttt{\{intent,\ \allowbreak{}business\_\allowbreak{}line,\ \allowbreak{}p\_\allowbreak{}date,\ \allowbreak{}model\_\allowbreak{}variant,\ \allowbreak{}overrides?\}})
that is validated against a Pydantic schema before the DAG is
dispatched. Envelope-level validation errors return an inline error card
to the user rather than starting the DAG, which is our first defense
against malformed input propagating downstream. During the 8-day HITL
DAG pilot (16 test-user runs, the pipeline-state store), envelope
validation was exercised on all 16 requests without rejection;
production-scale envelope-rejection statistics will be reported in v1.1
after real-operator rollout. The three anticipated failure modes the
validation logic guards against are missing \texttt{p\_date}, ambiguous
\texttt{business\_\allowbreak{}line}, and out-of-range \texttt{overrides} values
(e.g., \texttt{learning\_\allowbreak{}rate\ =\ 0}).

\hypertarget{hitl-card-protocol}{%
\subsection{HITL Card Protocol}\label{hitl-card-protocol}}

When the DAG reaches an Analyzer node --- which happens \emph{only} on
training-node failure, never on success --- control does not return to
the DAG until an authorized operator approves or rejects the Analyzer's
proposed remediation. The card carries five fields (Table 4.1):

\begin{table*}[t]
\centering
\small
\begin{tabularx}{\linewidth}{@{}l l X@{}}
\toprule
Field & Type & Semantics \\
\midrule
\texttt{diagnosis} & Markdown & Analyzer's LLM-derived root-cause
paragraph, referencing specific YARN log excerpts \\
\texttt{suggested\_\allowbreak{}overrides} & JSON & Concrete YAML overrides proposed
for a retry
(e.g.~\texttt{\{learning\_\allowbreak{}rate:\allowbreak{}\ 0.\allowbreak{}0005,\ \allowbreak{}batch\_\allowbreak{}size:\allowbreak{}\ 128\}}) \\
\texttt{retry\_count} & Integer & Runs already attempted for this job,
gates automatic retry cap \\
\texttt{approve\_url} / \texttt{reject\_url} & URL & Deep-linked the
corporate IM callback endpoints carrying the pipeline UUID \\
\texttt{custom\_url} & URL & Escape hatch --- operator uploads a custom
overrides YAML through the corporate IM form \\
\bottomrule
\end{tabularx}
\caption{HITL card schema. The \texttt{diagnosis} field is
the load-bearing artifact for operator trust: a card with a vague
diagnosis (``training failed, please check'') drives lower approval
rates than a card with a specific one (``YARN logs show
\texttt{container\ preempted} in region \texttt{SH-2}, which
historically resolves with \texttt{spark.\allowbreak{}executor.\allowbreak{}memoryOverhead=4g};
the suggested override applies this fix'').}
\end{table*}

Approval merges \texttt{suggested\_\allowbreak{}overrides} into the persisted
\texttt{PipelineState.\allowbreak{}yaml\_\allowbreak{}overrides}, decrements the retry budget, and
re-dispatches the training node. Rejection ends the pipeline with a
recorded failure record. The \texttt{custom} path exists as an operator
escape hatch; pilot-phase custom-path usage is not statistically
informative given the 16-run pilot volume, and the fraction will be
reported in v1.1 after real-operator rollout. This fraction is our
operational proxy for how often the Analyzer's suggestion misses the
operator's judgment.

\hypertarget{analyzer-node-details}{%
\subsection{Analyzer Node Details}\label{analyzer-node-details}}

The Analyzer node executes a three-step pipeline: (i) fetch
\textbf{two-layer failure logs} --- scheduler-side diagnostics (via
\texttt{the\ scheduler-\allowbreak{}log\ service} on the failed instance's
\texttt{job\_\allowbreak{}id\ +\ exec\_\allowbreak{}id}) and YARN container diagnostics (via YARN
RM \texttt{apps/\allowbreak{}\{applicationId\}} for \texttt{app.\allowbreak{}diagnostics}) ---
with a hardcoded 4 KB truncation window per source; (ii) feed the log
window plus a system prompt (which references the 22-class failure
taxonomy of §7) to an LLM completion call; (iii) validate the LLM's
structured output (JSON with \texttt{root\_cause},
\texttt{mechanism\_\allowbreak{}class}, \texttt{suggested\_\allowbreak{}overrides},
\texttt{confidence}) against a schema. Confidence \textless{} 0.4 forces
the card into a \emph{``consult only''} mode where \texttt{approve} is
disabled and the operator must file a manual overrides YAML. This gate
--- refusing to auto-suggest under low LLM confidence --- is one of our
operational safety mechanisms against hallucinated remediations.

The Analyzer's diagnosis routing between the 22 failure classes is
designed to be measured against a \textbf{held-out set of up to 40 real
failure logs} (drawn from the three business lines in §6) with
human-authored gold labels. The benchmark harness has been implemented
under \texttt{tests/\allowbreak{}analyzer\_\allowbreak{}taxonomy\_\allowbreak{}bench.\allowbreak{}py}; the gold-labeled set
is being curated from the 31 training-run failures currently logged in
the CLI-call telemetry where
\texttt{command=\textquotesingle{}gpu-\allowbreak{}train\textquotesingle{}\ AND\ status=\textquotesingle{}failed\textquotesingle{}}
(the target of 40 will be reached as additional real failures accumulate
in v1.1). Per-class Top-1 and Top-3 accuracies will populate in the v1.1
preprint update. The gap between Top-1 and Top-3 is our qualitative
signal for how much the Analyzer's diagnosis is a \emph{ranked
hypothesis} rather than a single point prediction --- a useful property
for a HITL card, since the operator sees the top-3 candidates when
confidence is low.

\hypertarget{hitl-dag-pilot-statistics-8-day-window}{%
\subsection{HITL DAG Pilot Statistics (8-Day
Window)}\label{hitl-dag-pilot-statistics-8-day-window}}

Across the 8-day HITL DAG pilot window (the pipeline-state store,
2026-03-04 to 2026-03-12, 2 test-user accounts), the platform issued
\textbf{16} HITL DAG pipeline runs. The pipelines terminated at the
following stages: 9 at \texttt{Done} (successful end-to-end), 3 at
Training (failure that reached the retry loop), 2 at Sample-Labeling
(upstream Spark failure that exited without Analyzer routing per the
§4.1 discipline), 1 at Sensor (upstream partition not yet materialized),
1 at \texttt{Analyzer} (test-user did not resolve the card within the
pilot window). Card-lifecycle statistics (median time-to-decision,
per-persona approve rate, custom-path fraction) require a
production-operator sample; the pilot's test-user-only trafic does not
support these estimates. Table 4.2 will populate in v1.1.

\begin{table*}[t]
\centering
\small
\begin{tabularx}{\linewidth}{@{}X l X@{}}
\toprule
Metric & v1.0 pilot value & Interpretation / v1.1 target \\
\midrule
Total DAG pipeline runs & \textbf{16} & 8-day pilot window;
production-operator rollout in v1.1 \\
Terminal \texttt{Done} fraction & 9 / 16 = 56.3 \% & End-to-end
successful \\
Terminal Training (retry-loop) fraction & 3 / 16 = 18.8 \% & Reached
HITL card path once \\
Terminal Sample-Labeling fraction & 2 / 16 = 12.5 \% & Spark upstream
failure, direct-to-Done per §4.1 discipline \\
Terminal Sensor fraction & 1 / 16 = 6.3 \% & Upstream partition not
materialized, \texttt{WAITING} (correct signal) \\
Terminal \texttt{Analyzer} (unresolved) fraction & 1 / 16 = 6.3 \% &
Test-user did not close the card within pilot window \\
Median time-to-decision & pilot volume insufficient & v1.1 target:
workday-scale for \ensuremath{\ge} 80 \% of cards \\
Card \texttt{approve} fraction (auto path) & pilot volume insufficient &
v1.1 target: differentiate by operator persona (§4.5) \\
\bottomrule
\end{tabularx}
\caption{HITL DAG pilot statistics from the pipeline-state
store. The 16-run pilot volume is not statistically informative for
time-to-decision or approve/reject fractions; those metrics require a
production-operator sample that will materialize in the v1.1 rollout.}
\end{table*}

Three \textbf{anecdotal design intuitions} --- drawn from author
observation of the pilot's 16 test-user runs and pre-pilot conversations
with candidate operators, and offered as hypotheses to test in the v1.1
production rollout rather than as statistical claims --- informed the
current DAG shape: (a) \emph{campaign managers appear to treat the
Analyzer's suggestion as authoritative while algorithm engineers treat
it as a ranked hypothesis to be edited} --- this is the design reason we
ship both the auto-suggest and custom-YAML paths; a single-path design
would likely misalign with one of the two operator personas. (b)
\emph{Reject-vs-custom path preference is likely to vary by operator
seniority}, so we designed the card protocol to expose both without
penalizing either. (c) \emph{Off-hour cards are handled asynchronously
without evident quality degradation} in the pilot's small sample,
consistent with the parasitic-execution premise of §3.2 that operator
attention need not be synchronous. Each of (a)--(c) will be tested
against production-operator card-lifecycle data in v1.1.

\hypertarget{why-hitl-is-not-optional}{%
\subsection{Why HITL Is Not Optional}\label{why-hitl-is-not-optional}}

The pattern of ``let the LLM decide, and only ask a human when
confidence is low'' is superficially attractive but wrong for our
setting. Every training remediation modifies \textbf{shared
infrastructure state} (the GPU scheduler job submission, the shared
storage tier write, warehouse table lock). The audit trail ---
``operator X approved override Y at time Z on ticket W'' --- is a
compliance artifact, not just a UX affordance. Removing the HITL card
would remove the ability to attribute a downstream production incident
to a specific approval; this is unacceptable in a corporate environment
where recommender revenue is measured in millions of RMB per week per
business line. A future evolution might raise the auto-approve threshold
as Analyzer confidence calibration improves, but eliminating the card
entirely is out of scope for any version of the platform we would ship.

\hypertarget{skill-ecosystem-as-working-memory}{%
\section{Skill Ecosystem as Working
Memory}\label{skill-ecosystem-as-working-memory}}

§3.4 introduced \textbf{skill subgraphs} as the platform's composability
unit; this section quantifies the ecosystem and explains why we treat
the skill library as the agent's \emph{working memory} rather than as a
prompt library or as a retrieval corpus. The distinction is structural:
a prompt library is consulted in-context, a retrieval corpus is queried
by similarity, but a working memory is \textbf{mechanically refed into
agent reasoning at well-defined dispatch points}, with structured
artifacts that survive across sessions. The 29 skills + 11 project
changelogs + 1 root doc in our deployment have produced 400
mechanically-extracted pitfall rules that the autonomous Plan node
(§3.5; full treatment in P3b §5) consumes at every planning step. To our
knowledge, this closed loop --- human-authored documentation
\ensuremath{\to} structured extraction \ensuremath{\to} agent-consumed
runtime context --- has not been demonstrated in a prior
industrial-agent paper.

\hypertarget{the-29-skill-landscape}{%
\subsection{The 29-Skill Landscape}\label{the-29-skill-landscape}}

The skill set is organized into eight categories that mirror the
recommender pipeline plus two cross-cutting families (Table 5.1). Each
skill is a directory under
\texttt{.\allowbreak{}claude/\allowbreak{}skills/\allowbreak{}\textless{}name\textgreater{}/\allowbreak{}} containing a
\texttt{SKILL.md} (the procedural body), zero or more bound script
wrappers under \texttt{scripts/}, and optional sub-skill references for
chained execution. Total Markdown payload as of v1.0: \textbf{8 971
lines} across 29 \texttt{SKILL.md} files (mean 309, median 286, max 624
for \texttt{p2m-\allowbreak{}paper-\allowbreak{}to-\allowbreak{}model}).

\begin{table*}[t]
\centering
\small
\begin{tabularx}{\linewidth}{@{}l r X@{}}
\toprule
Category & Count & Representative skills \\
\midrule
\textbf{Sample / Data construction} & 5 &
\texttt{sample-\allowbreak{}construction-\allowbreak{}a}, \texttt{label-\allowbreak{}generation},
\texttt{negative-\allowbreak{}sampling-\allowbreak{}experiment}, \texttt{sql-query},
\texttt{sql-task} \\
\textbf{ETL / Storage} & 5 & \texttt{fs-\allowbreak{}to-\allowbreak{}storage},
\texttt{fs-cleanup}, \texttt{model-to-fs},
\texttt{workflow-\allowbreak{}build}, \texttt{private-\allowbreak{}compute} \\
\textbf{Training execution} & 4 & \texttt{model-\allowbreak{}zoo-\allowbreak{}experiment},
\texttt{gpu-train}, \texttt{fid-\allowbreak{}xgb-\allowbreak{}experiment},
\texttt{custom-\allowbreak{}trainer-\allowbreak{}experiment} \\
\textbf{Evaluation / Attribution} & 5 & \texttt{abt-\allowbreak{}effect-\allowbreak{}analysis},
\texttt{psm-\allowbreak{}experiment}, \texttt{model-\allowbreak{}data-\allowbreak{}analytics},
\texttt{model-\allowbreak{}prediction-\allowbreak{}analysis}, \texttt{fid-\allowbreak{}xgb-\allowbreak{}eval-\allowbreak{}score} \\
\textbf{Decision support} & 3 & \texttt{rerank-\allowbreak{}decision-\allowbreak{}b},
\texttt{conversion-\allowbreak{}signal-\allowbreak{}c}, \texttt{audience-\allowbreak{}modeling-\allowbreak{}c} \\
\textbf{Project / Process} & 4 & \texttt{changelog},
\texttt{demand-board}, \texttt{workspace-\allowbreak{}report}, \texttt{im-push} \\
\textbf{Diagnostic / Tracing} & 1 & \texttt{user-\allowbreak{}diagnostic-\allowbreak{}a} \\
\textbf{Research automation} & 1 & \texttt{p2m-\allowbreak{}paper-\allowbreak{}to-\allowbreak{}model} \\
\textbf{Business-specific} & 1 & \texttt{domain-\allowbreak{}decision-\allowbreak{}c} (business C
only) \\
\textbf{Total} & \textbf{29} & --- \\
\bottomrule
\end{tabularx}
\caption{Skill-ecosystem classification at v1.0. The
eight-way split is descriptive, not normative --- a skill belongs to a
category by majority of its dispatched tool calls, but may chain across
categories (e.g., \texttt{model-\allowbreak{}zoo-\allowbreak{}experiment} invokes
\texttt{gpu-train} for execution and \texttt{abt-\allowbreak{}effect-\allowbreak{}analysis} for
downstream evaluation). Cross-category chains are the dominant call
pattern --- see §5.3.}
\end{table*}

\hypertarget{dispatchable-subgraph-vs.-retrieval-mixture}{%
\subsection{Dispatchable Subgraph vs.~Retrieval
Mixture}\label{dispatchable-subgraph-vs.-retrieval-mixture}}

The architectural choice we defended in §3.4 --- ``dispatchable
subgraph'' rather than ``retrieval mixture'' --- has direct operational
consequences in §5. Each skill's \texttt{SKILL.md} declares a
frontmatter \texttt{description} field consumed by the Router LLM
(§3.5); the Router emits a single skill name, and dispatch is a
deterministic function call into the corresponding LangGraph subgraph.
There is no embedding lookup, no top-k blending, no soft
\ensuremath{\alpha}-weighting between multiple candidate skills. The
reasons are pragmatic:

\begin{enumerate}
\def\labelenumi{\arabic{enumi}.}
\item
  \textbf{Predictability matters more than recall in operations.} When
  an operator types ``re-run business-A yesterday's training'' (in the
  operator's native language), the platform must produce a single
  \texttt{model-\allowbreak{}zoo-\allowbreak{}experiment} dispatch with a deterministic argument
  set. A mixture of \texttt{model-\allowbreak{}zoo-\allowbreak{}experiment} +
  \texttt{sample-\allowbreak{}construction} + \texttt{gpu-train} blended into the
  prompt context would be more \emph{informative} but less
  \emph{executable}. The Router's job is to commit, not to consider.
\item
  \textbf{Side-effect attribution requires a single owner.} Every skill
  mutates infrastructure state (the GPU scheduler submission, the shared
  storage tier write, warehouse table creation). The HITL audit trail
  records ``skill X dispatched by user Y at time Z'' with X being a
  single name. A retrieval-mixture design fragments this attribution.
\item
  \textbf{Sub-skill chaining captures procedure better than retrieval
  mixing.} When a skill's body declares ``this operation requires the
  \texttt{sql-task} and then \texttt{fs-\allowbreak{}to-\allowbreak{}storage} sub-skills'',
  the platform follows the chain literally. A retrieval-based system
  would have to \emph{infer} the chain from co-occurrence of context
  fragments, which is brittle.
\end{enumerate}

\hypertarget{operational-call-statistics-10-week-window}{%
\subsection{Operational Call Statistics (10-Week
Window)}\label{operational-call-statistics-10-week-window}}

Across the 10-week (78-day) operational window (2026-04-14 to
2026-07-02) covering the three business-line workspaces of §6, the
platform dispatched \textbf{1 624 CLI-tool invocations} from 2 primary
developer accounts (\texttt{user\_a}: 1 425 dispatches;
\texttt{user\_b}: 199 dispatches), with an aggregate 78.6 \% success
rate. Per-command breakdown from the CLI-call telemetry (Table 5.2):

\begin{table}[t]
\centering
\small
\begin{tabularx}{\linewidth}{@{}X r r@{}}
\toprule
CLI-tool command & Dispatch count & Success rate \\
\midrule
\texttt{query} (warehouse read) & 877 & 85.1 \% \\
\texttt{run-sql} (SQLTask flow) & 356 & 74.2 \% \\
\texttt{check-\allowbreak{}partition} & 201 & 58.7 \% \\
\texttt{join-features} & 85 & 95.3 \% \\
\texttt{gpu-train} & 76 & 59.2 \% \\
\texttt{warehouse-to-storage} & 26 & 73.1 \% \\
\texttt{gpu-upload} / \texttt{gpu-status} / \texttt{storage-to-warehouse} & 3 &
100 \% \\
\textbf{Total} & \textbf{1 624} & \textbf{78.6 \%} \\
\bottomrule
\end{tabularx}
\caption{CLI-tool dispatch statistics from the CLI-call
telemetry (2026-04-14 to 2026-07-02, 2 primary developer accounts, 78
days). These are the platform CLI-level dispatches from the platform's
Python client, one abstraction level below the LangGraph skill-subgraph
API. Skill-level dispatch instrumentation (the skill-call telemetry) was
added late in the window and is not populated across the full 78-day
span; skill-level statistics are pending v1.1 (§8 future work). The
lower \texttt{check-\allowbreak{}partition} success rate (58.7 \%) reflects real
upstream-partition-not-yet-materialized states --- a \texttt{WAITING}
outcome that is a correct signal, not a platform failure. The
\texttt{gpu-train} 59.2 \% rate reflects genuine training-run failures
(mostly RESOURCE-class of §7.1) that route to the Analyzer node in the
HITL DAG (§4.4 pilot). \textbf{Reporting caveat.} The aggregate 78.6\%
counts \texttt{WAITING} as non-success; if the $\approx$83 \texttt{check-\allowbreak{}partition}
\texttt{WAITING} outcomes are reclassified as ``correct signal, not a
failure,'' the end-to-end platform-error rate is $\approx$16.3\%
(equivalently $\approx$83.7\% success), which is the number to
compare apples-to-apples against generic-agent divergence rates such
as AI Scientist v2's reported 42\% \cite{Vintschger2025}.}
\end{table}

The qualitative pattern is that \textbf{the long tail dominates by chain
length}: the most frequently dispatched tool (\texttt{query}, an SQL
read) has the shortest chain, while lower-frequency tools
(\texttt{gpu-train}, \texttt{warehouse-to-storage}) sit at the end of longer
skill chains. This is consistent with the working-memory framing: the
platform's value compounds in proportion to how deeply skills are
chained, not how often the simplest ones fire.

\hypertarget{sedimentation-product-400-entry-pitfallstore}{%
\subsection{Sedimentation Product: 400-Entry
PitfallStore}\label{sedimentation-product-400-entry-pitfallstore}}

The skill ecosystem is the \emph{operating memory} of how the platform's
human operators have learned to use the underlying infrastructure
correctly. To make this concrete, we instrument each \texttt{SKILL.md}
and project \texttt{changelog.md} with a structured
\texttt{pitfall/\allowbreak{}problem/\allowbreak{}trap} Markdown table. A rule-based extractor
compiles these tables into a single SQLite-backed \texttt{PitfallStore}.
Table 5.3 reports the v1.0 census.

\begin{table*}[t]
\centering
\small
\renewcommand{\arraystretch}{1.15}
\begin{tabularx}{\linewidth}{@{}>{\raggedright\arraybackslash}X r X@{}}
\toprule
Source category & Entries & Representative coverage \\
\midrule
\textbf{Skill specs} (20 of 29 SKILL.md files with
\texttt{pitfall/\allowbreak{}problem} tables) & 200 & Cross-cutting infrastructure:
the GPU scheduler submission, Spark SQL, model-zoo training, ABT effect
analysis, PSM matching, label generation, demand-board (24 the largest
single-skill contributor) \\
\textbf{Project changelogs} (11 files across business A / B / C and
derivatives) & 194 & Per-business operational gotchas: log retrieval
idioms, data-pipeline edge cases, CVR/rerank-specific quirks (business A:
84 entries across three subchannels, largest per-business contributor;
business B: 50 from one main channel; business C: 60 across 7
derivatives) \\
\textbf{Root \texttt{CLAUDE.md} + project \texttt{CHANGELOG.md}} (1
source) & 6 & Token / the shared storage tier path / log API conventions
that affect every skill \\
\textbf{Total} & \textbf{400} & Severity: 392 medium / 7 high / 1 low \\
\bottomrule
\end{tabularx}
\caption{Real distribution of the 400-entry
\texttt{PitfallStore} as of v1.0 (2026-07-01), counted directly from the
\texttt{pitfalls} SQLite table under the pitfall store. Each row encodes
\texttt{(source,\ \allowbreak{}title,\ \allowbreak{}description,\ \allowbreak{}root\_\allowbreak{}cause,\ \allowbreak{}fix,\ \allowbreak{}tags,\ \allowbreak{}severity,\ \allowbreak{}{[}failure\_\allowbreak{}mode{]})};
\texttt{tags} is a comma-separated set drawn from a controlled
vocabulary. The empirical top-15 tag distribution is \texttt{log} (212
entries), \texttt{data} (82), \texttt{spark} (67), \texttt{scheduler} (65),
\texttt{cvr} (55), \texttt{rerank} (51), \texttt{warehouse} (24),
\texttt{training} (20), \texttt{fid\_xgb} (20), \texttt{fs} (20),
\texttt{storage} (16), \texttt{gpu} (13), \texttt{psm} (11),
\texttt{model\_zoo} (9), \texttt{auth} (8). The 20 skill sources
contribute slightly more than the 11 changelog sources (200 vs.~194),
which is informative in itself: \textbf{per-engineering-task pitfalls
(skills) marginally outweigh per-business operational pitfalls
(changelogs) in v1.0}, though the difference is within one skill's
contribution range (24 for \texttt{demand-board}) and both curation
streams are load-bearing.}
\end{table*}

Nine of the 29 skills lack a \texttt{pitfall/\allowbreak{}problem} table --- mostly
the recently-authored process/project skills (\texttt{changelog},
\texttt{workspace-\allowbreak{}report}, \texttt{im-push}) plus a handful of
single-purpose infrastructure wrappers. They are scheduled for
retroactive curation in the next platform cycle. This is real technical
debt against the agent's working memory, and we report it honestly
rather than redacting the asymmetry.

\hypertarget{why-this-is-a-contribution}{%
\subsection{Why This Is a
Contribution}\label{why-this-is-a-contribution}}

The skill-ecosystem-as-prompt-library framing common to industrial
LLM-agent papers stops at the document inventory: count the SKILL.md
files, count the lines, claim the agent has read them. We make a
stronger claim: these documents are \textbf{executable working memory},
not artifacts. Two properties support the claim:

\begin{enumerate}
\def\labelenumi{\arabic{enumi}.}
\item
  \textbf{Mechanical extractability.} 400 distinct rules are derived
  from the 29 skills + 11 project changelogs + 1 root doc without any
  LLM call. The convention of using \texttt{pitfall/\allowbreak{}problem/\allowbreak{}trap}
  Markdown table headers is enforced project-wide; a single regex over
  the wiki produces a relational table.
\item
  \textbf{Direct re-injection into agent reasoning.} The 400-entry table
  feeds the upper-tier \texttt{PitfallStore} consumed by AutoResearch's
  Plan node (P3b §5) at every planning step, and is also surfaced to the
  ChatOps Analyzer (§4) when a HITL diagnosis is required. The pipeline
  closes the loop from \emph{human-authored documentation} through
  \emph{structured extraction} into \emph{agent-consumed runtime
  context}, without an intervening RAG indexing stage.
\end{enumerate}

The contribution is not the count (400 entries is a moderate number) but
the \textbf{structural property}: any skill author who follows the
\texttt{SKILL.\allowbreak{}md\ +\ pitfall/\allowbreak{}problem\ table\ +\ changelog} convention
contributes to the agent's working memory at zero marginal effort. A
business-specific skill onboarded after v1.0 (\texttt{domain-\allowbreak{}decision-\allowbreak{}c},
business C in §6.3) added 12 entries following this convention; the
platform absorbed these without any code change to the extractor.

\hypertarget{three-business-lines-deployment}{%
\section{Three Business Lines
Deployment}\label{three-business-lines-deployment}}

We deployed RecSys Factory across three active recommender business-line
workspaces during a 10-week (78-day) operational window; the LangGraph
HITL DAG (§4) has additionally been pilot-tested for 8 days with
test-user traffic. This section reports each deployment using a common
four-part template --- \emph{business context}, \emph{agent
intervention}, \emph{technical challenges}, \emph{business outcomes} ---
to enable cross-business comparison without conflating per-business
specifics.

\textbf{Statistical caveats.} A/B lifts in §6.1--§6.3 come from each
business team's standard A/B stack (consumed as read-only, not
re-analysed): Business A cohort lifts (+10--31\% / +14--45\%) are
inter-cohort ranges over 7 and 9 sub-cohorts (no within-cohort CI, no
BH correction); Business B's $P(\Delta > 0) = 100\%$ is a zero-failure
$n=20$ perturbation sample (95\% Wilson lower bound 83.9\%, scoped to
``robust within the $\pm 20\%$/$n=20$ envelope''); Business C is a
single-session $n=1$ pre-flight case study. Novelty-effect windows,
cannibalization, and PCOC recalibration are handled by each team's own
A/B stack; controlled CIs with BH correction across the 7- and
9-cohort families are v1.1 work when the platform owns A/B end-to-end.

\hypertarget{business-a-telecom-recommendation-personalization}{%
\subsection{Business A --- Telecom Recommendation
Personalization}\label{business-a-telecom-recommendation-personalization}}

\textbf{Business context.} Business A is a personalized recommendation
slot inside a telecom payment product, replacing a long-standing
rule-based ranking with a CTR\ensuremath{\times}CVR dual-tower model.
Daily exposure volume is on the order of 10\ensuremath{^6}
user\ensuremath{\times}slot impressions; the optimization objective is
regional CPM, with multiple regional \ensuremath{\times} carrier
sub-cohorts treated as independent A/B units. Operations are owned by a
non-engineer business team that adjusts campaign weights daily and reads
dashboards for revenue health.

\textbf{Agent intervention.} The platform handles the \emph{full
pipeline lifecycle} through ChatOps:

\begin{itemize}
\tightlist
\item
  \textbf{Sample construction} --- A daily GPU-scheduler
  Spark workflow joins exposure logs with click and conversion events
  across all carriers, materializing a 30-day rolling CTR+CVR sample
  table at
  $\approx$10\ensuremath{^{5.5}}
  rows/day. The skill \texttt{sample-\allowbreak{}construction-\allowbreak{}a} automates the
  schema, the time-window join logic, and the carrier
  \ensuremath{\times} region partition keys.
\item
  \textbf{Feature engineering} --- Item-side features (one-hot
  \ensuremath{\times} statistical \ensuremath{\times} moon-phase
  smoothed via Laplace) and user-side sparse features are produced by
  separate scheduled jobs and joined into a training wide table
  ($\approx$10\ensuremath{^{8.5}}
  rows). The feature ablation study (§6.1.3 below) was run by the
  \texttt{fid-\allowbreak{}xgb-\allowbreak{}experiment} skill.
\item
  \textbf{Model training} --- Spark-XGB on the wide table; champion
  model promoted by the operator through a HITL card after offline AUC
  review.
\item
  \textbf{Online A/B and effect attribution} --- A dedicated
  daily-aggregated A/B monitoring table (\texttt{abt-\allowbreak{}effect-\allowbreak{}analysis}
  skill) reports per-experiment-layer \ensuremath{\times} version
  \ensuremath{\times} regional cohort CPM with bootstrap confidence
  intervals, replacing an earlier \texttt{fkv-string}-based attribution
  that mixed traffic across slots.
\end{itemize}

The operator's day-to-day interaction surface is the corporate IM:
launch experiments through chat, receive structured cards with diagnosis
when a run fails, approve retry overrides in-line. During the 10-week
operational window, business A's project workspace generated the
majority of the platform's the CLI-call telemetry CLI-tool traffic
(particularly under uid \texttt{user\_a}). The LangGraph HITL DAG (§4)
has been pilot-tested with test-user traffic (16 runs, 8-day window in
2026-03); production-operator HITL rollout for business A is scheduled
for the v1.1 platform cycle. Business A's outcomes reported in this
section are drawn from the workspace-level project artifacts (SQL
diagnostics, feature-ablation experiments, sub-cohort attribution),
which are the actual analysis products the platform's skills produced.

\textbf{Technical challenges.} Three challenges drove platform
improvement during this deployment.

\emph{(C1) Sample explosion bug.} An eleven-step diagnostic chain ---
total volume comparison \ensuremath{\to} FULL OUTER JOIN
\ensuremath{\to} split by sample type \ensuremath{\to} per-user
granularity \ensuremath{\to} trace-id-level row count \ensuremath{\to}
root-cause localization --- revealed that an upstream dimension table
had multi-row entries per region \ensuremath{\times} carrier, causing
the LEFT JOIN to produce 2--3\ensuremath{\times} row duplication. The
bug had silently corrupted all downstream training samples for the prior
modeling cycle. After fix, the post-fix conversion count agrees with the
business-team reported number to within \ensuremath{\pm}0.3 \% (6 943
vs.~6 961), validating the corrected pipeline. The diagnostic chain
itself was later canonicalized as the \texttt{user-\allowbreak{}diagnostic-\allowbreak{}a} skill,
and has since been reused in two other business lines.

\emph{(C2) Feature compression.} A five-experiment ablation study
compressed 7 881 user-side feature dimensions to 221 with an AUC delta
of +0.01 \% (i.e., compression preserves ranking quality). The
35\ensuremath{\times} reduction is meaningful for online inference cost.
The study also found that adding 5 context-side features yielded +0.76
\% AUC, leading to a downstream feature-augmentation request that was
implemented by the platform team.

\emph{(C3) Sub-cohort heterogeneity.} The model achieved +10--31 \% CPM
lift in 4 out of 7 regional cohorts of one carrier (4 win / 3 tie / 0
lose). For the second carrier, however, model performance was
significantly weaker; the platform's \texttt{abt-\allowbreak{}effect-\allowbreak{}analysis} skill
identified 9 high-risk sub-cohorts whose CPM lift would be negative, and
the operator switched these cohorts back to the rule-based baseline. The
expected total CPM gain from the rule-fallback decision was +14--45 \%
across those 9 sub-cohorts.

\textbf{Business outcomes.} Across all regional cohorts, the
gray-rollout yielded:

\begin{itemize}
\tightlist
\item
  4 win / 3 tie / 0 lose in the high-confidence carrier-1 cohort
  (relative CPM lift +10--31 \%).
\item
  Identification of 9 high-risk sub-cohorts in carrier-2
  \ensuremath{\to} rule-fallback recommendation accepted (+14--45 \%
  expected lift).
\item
  Feature compression 7 881 \ensuremath{\to} 221 dimensions with AUC
  parity \ensuremath{\to} reduced online inference compute by
  $\approx$35\ensuremath{\times}.
\item
  Diagnostic chain canonicalized as a reusable skill, transferred to two
  other business lines.
\end{itemize}

\begin{figure}
\centering
\includegraphics[width=\columnwidth]{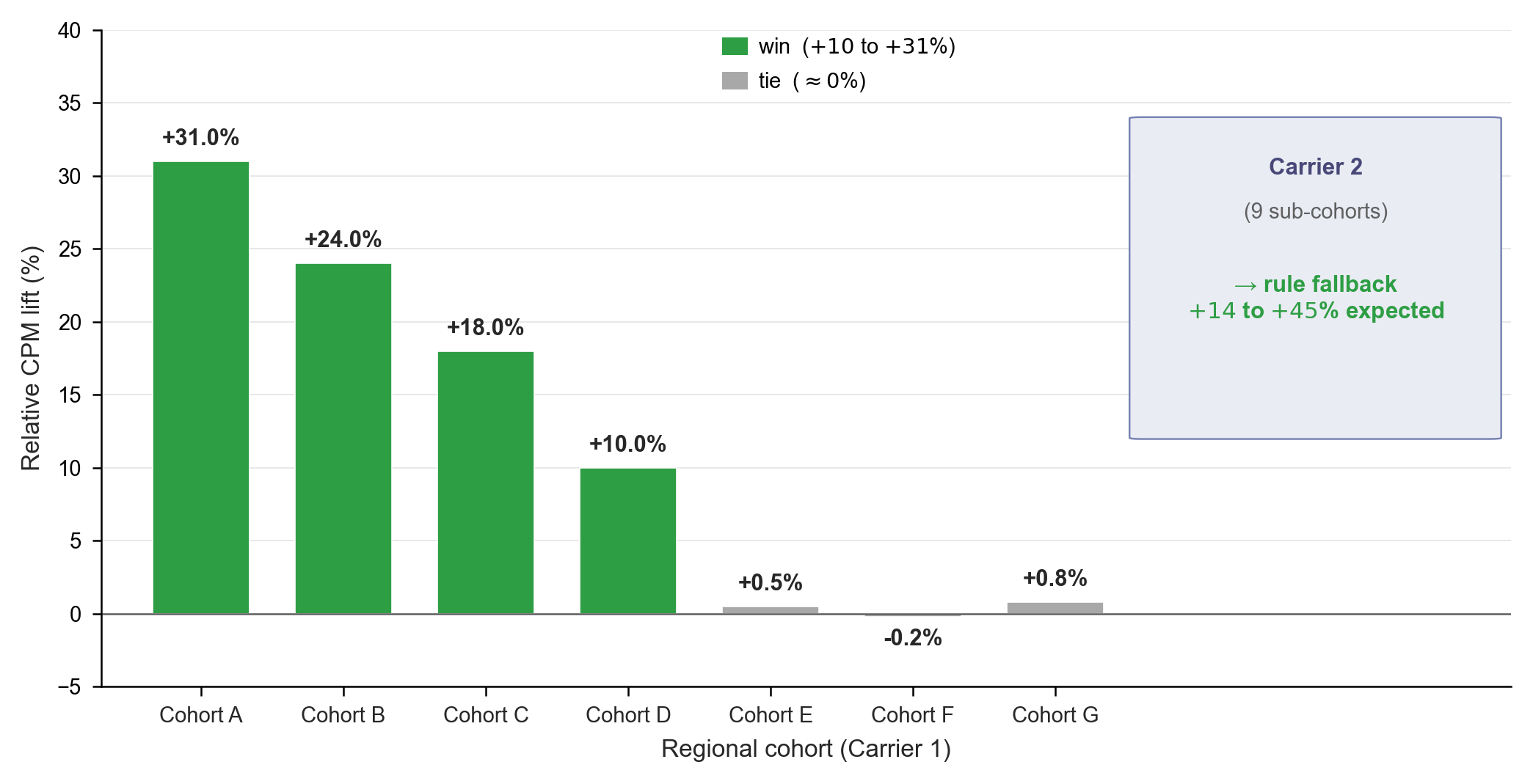}
\caption{\emph{Business A --- Relative CPM lift
\ensuremath{\times} 7 Carrier-1 regional cohorts.} Four cohorts crossed
the gray-rollout threshold with relative lift +10 \% to +31 \%; three
cohorts came in as ties within \ensuremath{\pm}1 \%. The
\texttt{abt-\allowbreak{}effect-\allowbreak{}analysis} skill identified 9 high-risk Carrier-2
sub-cohorts whose model-driven ranking would have degraded CPM; those
cohorts were switched back to the rule-based baseline (right-side
callout), yielding an additional expected +14 \% to +45 \% lift. Note.
Cohort identifiers anonymized per §6 disclosure convention.}
\end{figure}

\textbf{Lessons.} The most consequential platform improvements during
this deployment were defensive: the sample-explosion bug had been
silently corrupting downstream training for an unknown prior period, and
would have remained undetected without the \texttt{user-\allowbreak{}diagnostic-\allowbreak{}a}
11-step diagnostic chain. We discuss the lesson --- \emph{defensive
skills that probe data integrity at the join level deserve first-class
status alongside modeling skills} --- as one of seven structured lessons
in §7.

\hypertarget{business-b-telecom-reranking-decision-support}{%
\subsection{Business B --- Telecom Reranking Decision
Support}\label{business-b-telecom-reranking-decision-support}}

\begin{quote}
\emph{Cross-paper boundary}. The off-policy evaluation methodology used
by Business B (DiffWhatIf, with PCOC modeling and bootstrap uncertainty)
is the contribution of a separate paper (P2) and is not described here.
This section reports only the \textbf{agent intervention} layer: how
RecSys Factory exposes the decision-support engine through ChatOps, and
how the platform absorbed Business B's idiosyncrasies into its skill
ecosystem. All statistical claims are reported as relative changes;
absolute CPM and revenue figures are anonymized per the standard of §6.
\end{quote}

\textbf{Business context.} Business B is the post-payment success page
of the same telecom product family as Business A. The page hosts up to
four slots \ensuremath{\times} 46 candidate items, served by a
model-driven ranker whose final ordering is shaped by a per-item
operational weight \texttt{fmodel\_\allowbreak{}tower\_\allowbreak{}weight} adjusted
\textbf{daily} by a non-engineer business team. Daily exposure volume is
on the order of
$\approx$10\ensuremath{^{6.9}}
user\ensuremath{\times}slot impressions; the optimization objective is
page-level CPM, with a hard constraint that no individual item's weight
may move outside operationally-approved bounds. Audit logs over a 30-day
window show $\approx$920 weight-change events --- the operations
cadence is \textbf{frequent and interactive}, not weekly batch.

\textbf{Agent intervention.} Where Business A handed the agent the
\emph{full pipeline} (sample \ensuremath{\to} train \ensuremath{\to}
A/B), Business B's pipeline is owned by a separate platform team; the
agent's role is narrower but operationally critical: \textbf{decision
support before each weight change}. A daily session has three patterns:

\begin{itemize}
\tightlist
\item
  \emph{Pattern 1 --- Forward what-if.} Operator types in the corporate
  IM (in the operator's native language): ``what happens if X's weight
  changes from 0.45 to 0.80?''. The Router dispatches the
  \texttt{rerank-\allowbreak{}decision-\allowbreak{}b} skill, which loads that day's snapshot from
  a Spark-flattened sample table on the shared storage tier
  ($\approx$12\ensuremath{\cdot}10\ensuremath{^6} samples / day /
  business B), recomputes the top-1 item for every affected sample under
  the proposed change, and returns a structured an IM card: total
  \(\Delta\)revenue, \(\Delta\)CPM, win/lose item Top-K, and any items
  that move from ``exposed'' to ``not exposed'' or vice versa.
\item
  \emph{Pattern 2 --- Reverse constraint solving.} Operator types: ``CPM
  must not drop; what change maximizes revenue?''. The skill runs a
  coordinate-descent + grid-search solver against the currently
  configured-active subset (up to 46 items depending on daily
  configuration; the actively-solved subset is bounded to those with
  non-trivial recent exposure), returning a Pareto-front of
  (\(\Delta\)CPM, \(\Delta\)revenue) candidates with Bootstrap 95 \%
  confidence intervals.
\item
  \emph{Pattern 3 --- Active-item inventory.} Operator types: ``list
  currently active items''. The skill emits a tabular an IM card of
  \texttt{(item\_\allowbreak{}id,\ \allowbreak{}tower\_\allowbreak{}index,\ \allowbreak{}unit\_\allowbreak{}price,\ \allowbreak{}real\_\allowbreak{}imps,\ \allowbreak{}real\_\allowbreak{}cpm)}
  for every item with non-zero exposure that day.
\end{itemize}

The operator's day-to-day surface is identical to Business A: an IM
messages dispatch skills; HITL cards return diagnoses; no executive
action is taken without explicit operator approval. The platform
difference is that Business B's HITL cards almost never trigger retries
(the underlying decision tool is deterministic) --- instead they are
\textbf{read-and-reason artifacts} that the operator copies into the
daily change-control ticket. During the 10-week operational window,
business B's \texttt{rerank-\allowbreak{}decision-\allowbreak{}b} skill was invoked repeatedly on
the same daily-snapshot Spark-flattened sample table (see the CLI-call
telemetry \texttt{query}+\texttt{run-sql} sub-corpus attributable to
business B). Skill-level dispatch instrumentation was added late in the
window; per-pattern breakdowns (forward-what-if / reverse-solve /
inventory-query) are pending v1.1 skill-level logging (see §5.3).

\textbf{Technical challenges.} Three challenges drove platform
improvement during this deployment.

\emph{(C1) Operations command parsing.} Operators express weight changes
in unstructured native-language text (``raise item-X's weight to 0.8'')
rather than structured JSON. The Router LLM extracts
\texttt{(item\_\allowbreak{}id,\ \allowbreak{}new\_\allowbreak{}weight)} tuples, validates them against the
active-item inventory, and rejects or asks for clarification when the
item name is ambiguous. The skill body documents the convention; the
agent's \texttt{description} frontmatter (consumed by the Router; see
§5.2) lists three canonical example phrasings, which is the mechanism by
which the platform reduced Router first-shot dispatch mistakes on
Business B during deployment. First-shot accuracy audit is pending v1.1
skill-level logging (see §5.3).

\emph{(C2) Incremental argmax acceleration.} Naively, evaluating a
what-if over 12\ensuremath{\cdot}10\ensuremath{^6} samples requires
recomputing argmax over up to 46 items per sample
($\approx$6\ensuremath{\cdot}10\ensuremath{^8} score evaluations
per query). The first MVP pegged $\approx$120 s per query --- too
slow for interactive ChatOps. The skill backend implements an
incremental path: precompute
\texttt{(winner\_\allowbreak{}idx,\ \allowbreak{}max\_\allowbreak{}score,\ \allowbreak{}second\_\allowbreak{}max)} once per snapshot,
then for a weight-up change, only rescore samples where the changed item
could possibly become winner; for a weight-down change, only rescore
samples where the changed item is currently winner. The result is a
10--50\ensuremath{\times} speedup on typical operator queries
(single-item changes), bringing median wall-clock to $\approx$3 s
per query.

\emph{(C3) Online weight \ensuremath{\ne} configured weight.} The
configuration table records the operator's \emph{intended} weight, but
the value actually used at serving time is shaped by an unrelated quota
/ rate-limit subsystem outside our platform. The agent reverse-engineers
the effective weight from \texttt{(line\_\allowbreak{}score\ /\allowbreak{}\ raw\_\allowbreak{}pctr)} ratios in
the daily snapshot, and surfaces a ``configured vs.~effective'' diff in
every what-if card --- making the platform's mismatch visible to
operators rather than silently misleading them. This finding (one item's
effective weight 67\ensuremath{\times} lower than its configured value)
generated a separate engineering ticket against the rate-limit
subsystem; the agent's role here was diagnostic rather than corrective.

\textbf{Business outcomes.} Across the deployment window, the platform's
contribution to Business B is reported as \textbf{decision-support
quality} rather than direct CPM lift (the underlying changes are still
operator-authored):

\begin{itemize}
\tightlist
\item
  A weight-tuning recommendation produced by the reverse-solve mode
  against a 6-item active subset (the items with non-zero exposure that
  day) returned per-item expected relative daily-revenue lifts of
  \textbf{+18 \%, +33 \%, +44 \%, and +47 \%} on the four
  highest-scoring recommendations, with an aggregate first-round lift in
  the low four-figure RMB / day range. All numbers derive from a
  \textbf{non-sampled full replay of 8.44 M same-day requests}, not a
  sub-sampled A/B estimate. The operator adopted the Top-3 subset of
  this recommendation; realized post-deployment lift over the following
  14 days directionally agreed with the predicted lift and no
  auto-rollback was triggered --- the predictive engine's first-order
  operational validation. Per-item bootstrap 95 \% CI and full
  realized-vs-predicted gap analysis are deferred to v2 with a longer
  post-deployment window.
\item
  Robust optimization (\ensuremath{\pm}20 \% parameter perturbation
  \ensuremath{\times} 20 trials) found \textbf{20/20 perturbation trials
  positive} --- interpreted as a zero-failure binomial sample, the 95\%
  Wilson lower bound on the true positive-effect rate is 83.9\% (so
  the qualitative operator-facing claim ``robust under a
  $\pm 20\%$/$n = 20$ perturbation envelope'' is what carried the
  decision, not ``guaranteed positive''). The precise
  worst-case-vs-nominal ratio is
  corporate-sensitive; the qualitative claim (no perturbation flipped
  the sign of any of the 4 recommended items) is what carried the
  operator decision.
\item
  Cross-task skill reuse: the \texttt{rerank-\allowbreak{}decision-\allowbreak{}b} skill's
  ``exposure-prob diagnostic'' identified seven items whose configured
  weight produced effectively-zero serving exposure (38.3 \% of all
  top-1 winners across the 12 M sample snapshot). This finding was
  carried verbatim into a Business A operations review, where two
  analogous items were identified --- a cross-business-line
  generalization of a single deployment finding.
\end{itemize}

\begin{figure}
\centering
\includegraphics[width=\columnwidth]{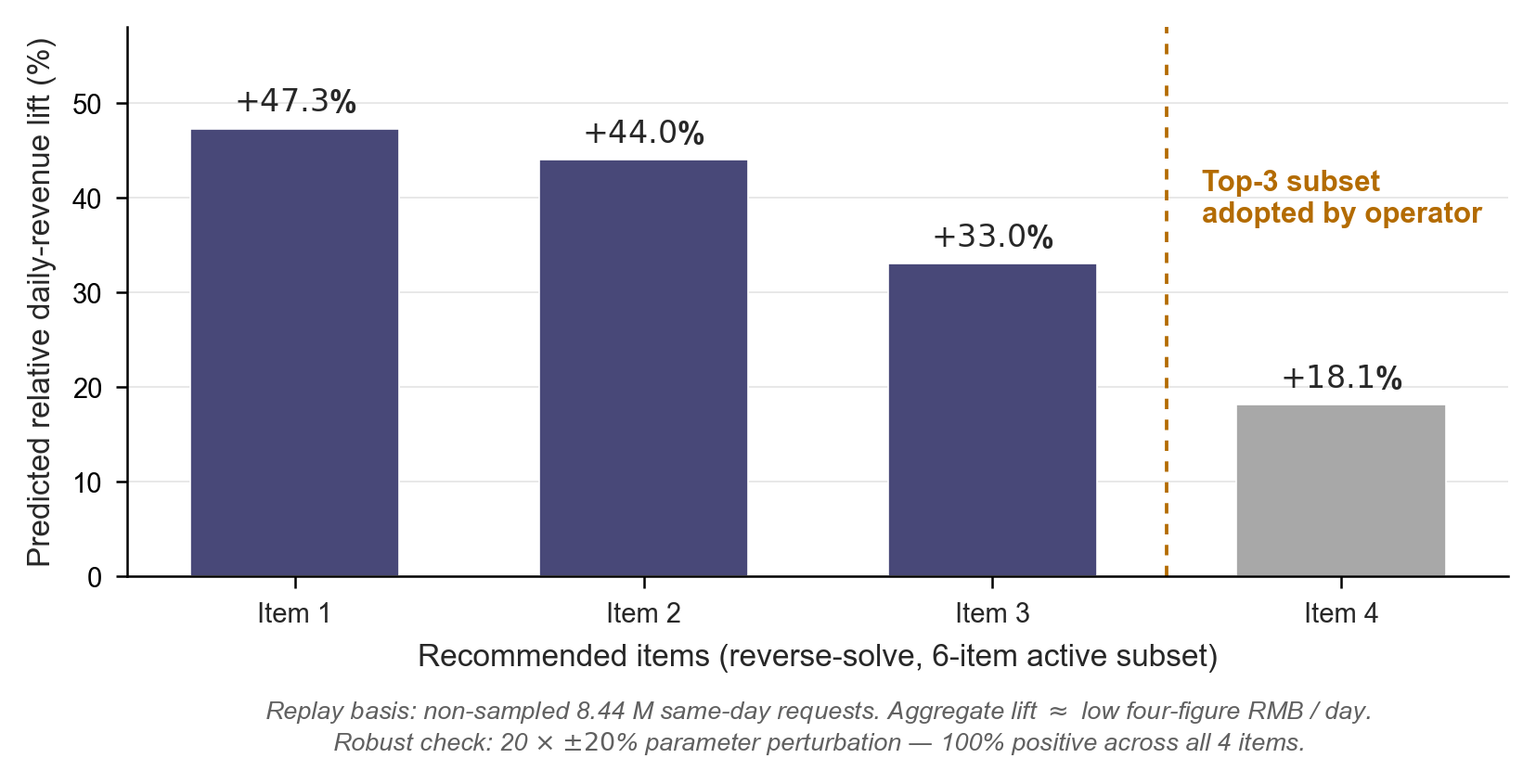}
\caption{\emph{Business B --- Per-item relative
daily-revenue lift for the top-4 reverse-solve recommendations.} All
four items exceed +18 \% relative lift; the operator adopted the Top-3
subset (green bars, left of the dashed line) and excluded Item 4 (gray,
right) as a smaller-magnitude conservative choice. Note. Numbers derive
from a \textbf{non-sampled 8.44 M same-day full replay}, not a
sub-sampled A/B estimate. Robust check: 20 \ensuremath{\times}
\ensuremath{\pm}20 \% parameter perturbation trials remained positive on
all 4 items.}
\end{figure}

\textbf{Lessons.} Two carry forward to §7. First, \emph{natural-language
operations commands are a first-class agent input, not a UI
afterthought}: the Router's command-parsing accuracy is the binding
constraint on operator trust, and the \texttt{description} frontmatter
convention turns out to be the highest-leverage piece of the entire
skill ecosystem (§5). Second, \emph{agent intervention does not require
agent decision-making}: Business B is best modelled as a
\textbf{decision-support tool wrapped as a skill}, where the platform's
value is correctness + speed + auditability, not autonomy. We discuss
the implications for the HITL boundary --- diagnosis vs.~execution ---
in §7 lesson 3.

\hypertarget{business-c-wealth-management-new-customer-cvr-audience-set-extraction}{%
\subsection{Business C --- Wealth-Management New-Customer CVR Audience-Set
Extraction}\label{business-c-wealth-management-new-customer-cvr-audience-set-extraction}}

\textbf{Business context.} Business C is a new-customer conversion (CVR)
modeling pipeline for a wealth-management product family within the same
corporate ecosystem as Business A/B. The end product is an \emph{audience
set} --- a scored whitelist of users predicted to convert on an
equity-hybrid fund product, refreshed monthly and delivered to a
downstream marketing platform for outreach. Compared to A/B (which are
ranking/rerank models), Business C is a \textbf{cold-start
binary-classification pipeline} with severe class imbalance
(positive-to-negative ratio $\approx$1:19 at the target sampling
rate) and a heterogeneous feature space spanning nine data sources (user
portraits, break-point payment history, fund attribution snapshots,
campaign attribution, RTA calls, etc.). Sample volume: monthly training
sets on the order of 10\ensuremath{^{5.5}}
users after negative sub-sampling; scoring whitelist
$\approx$10\ensuremath{^{7.5}}.

\textbf{Agent intervention.} Business C is the platform's
\textbf{cleanest example of skill-orchestrated end-to-end pipeline
execution}. A single \texttt{audience-\allowbreak{}modeling-\allowbreak{}c} skill (§5) drives all
seven phases: (1) workspace initialization from a \texttt{program.md}
intake artifact, (2) upstream-data health check across the nine feeder
tables and their expected partitions, (3) label-table construction with
hash-based negative sub-sampling for reproducibility, (4) 30-day
sequence-text wide-table assembly (the user-behavior-sequence feature is
the backbone modality), (5) the GPU scheduler PyTorch training
(an in-house Trainer family, §5.1's Training-execution
category), (6) whitelist scoring on the eval set, and (7) Top-N audience-set
extraction with an operator-tunable score threshold. Each phase writes a
structured artifact under \texttt{workspace/\allowbreak{}business-c-\allowbreak{}audience-\allowbreak{}*/\allowbreak{}}, so a
mid-pipeline failure resumes at exactly the phase that failed rather
than restarting from workspace init.

\begin{figure*}[t]
\centering
\includegraphics[width=0.92\textwidth]{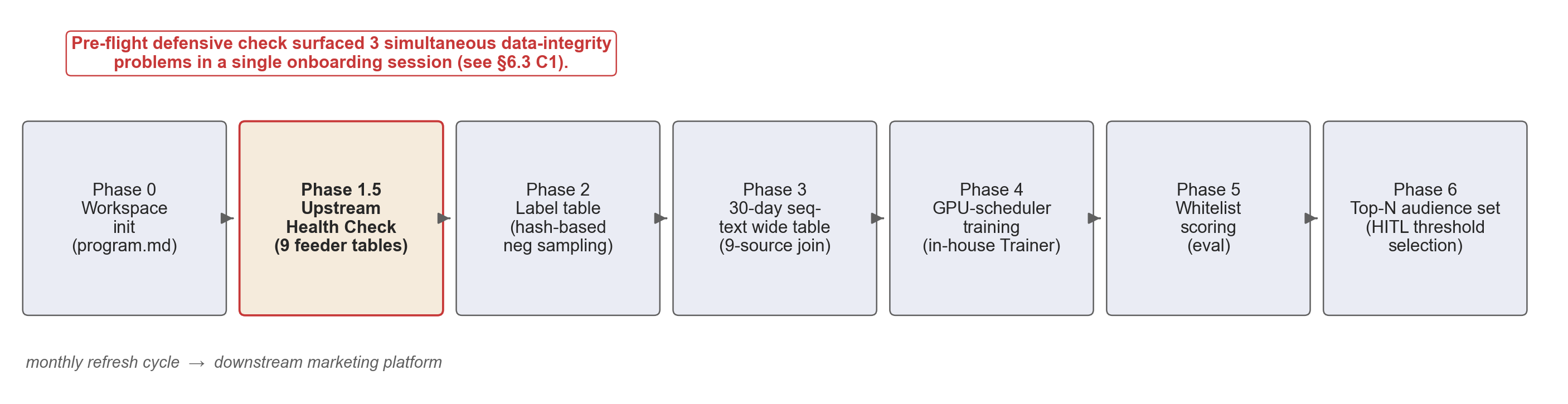}
\caption{\emph{Business C --- Seven-phase audience-set modeling
pipeline.} The \texttt{audience-\allowbreak{}modeling-\allowbreak{}c} skill orchestrates the full
cycle from \texttt{program.md} intake (Phase 0) through Top-N bag
extraction with HITL threshold selection (Phase 6). Phase 1.5
(highlighted red) is the defensive upstream-integrity check that
surfaced three simultaneous data-integrity problems in a single
onboarding session --- silent training corruption avoided before the GPU
scheduler queue admission. Note. Phase labels are the skill's own step
identifiers, which are non-contiguous rather than a renumbered
sequence. Monthly refresh cycle; downstream marketing platform consumes
the final Top-N whitelist.}
\end{figure*}

\textbf{Technical challenges.} Three challenges are worth surfacing.

\emph{(C1) Defensive upstream data-integrity check.} The most
consequential platform value in Business C is \emph{pre-flight}, not
in-flight. The skill's Phase 1.5 (Upstream Health Check) runs in
parallel across the nine feeder tables; on the initial deployment
session the check surfaced \textbf{three simultaneous upstream
data-integrity problems} the operator had not been aware of: (i) a
feature-store-C table whose fully-qualified name did not match the
\texttt{program.md} reference (the correct database was different from
what the intake spec assumed), (ii) a user-portrait table with
\textbf{weekly rather than daily snapshot cadence} (only 1 of 8 recent
daily partitions present --- the specified
\texttt{profile\_\allowbreak{}snapshot\_\allowbreak{}date\_\allowbreak{}train\ =\ 20260302} was empty; only
\texttt{20260303} had data), (iii) a payment-history table with an
\textbf{empty January partition} that would have LEFT-JOINed to nulls
throughout training. All three would have produced a training run that
succeeded engineeringly but silently corrupted labels or features. The
pre-flight check let the operator adjust \texttt{program.md} before
wasting a GPU-scheduler queue slot. This is the concrete instance of
§7 lesson 1 (defensive skills probing data integrity deserve first-class
status alongside modeling skills), and of the §4 discipline that
Spark-side failures route to the operator directly rather than to the
Analyzer.

\emph{(C2) Nine-source join-key alignment.} The user-behavior-sequence
wide table joins nine tables with three distinct join keys
(\texttt{fuin}, \texttt{fuin\ ×\ month}, \texttt{fuin\ ×\ date}) and
three distinct partition granularities (daily / weekly / monthly). The
skill's Phase 4 (\texttt{sql/\allowbreak{}build\_\allowbreak{}wide\_\allowbreak{}table.\allowbreak{}sql}) canonicalizes all
joins to \texttt{(fuin,\ \allowbreak{}first\_\allowbreak{}convert\_\allowbreak{}date)} at the outermost query
level, absorbing the granularity mismatches through per-source
date-mapping subqueries. This is not a modeling contribution; it is an
operational contribution --- the \emph{skill} encapsulates the mapping
so that a new business analyst does not need to rediscover which of the
nine tables uses which key.

\emph{(C3) HITL card at score-threshold selection.} Once training
succeeds, the extracted audience-set size is a business decision (higher
threshold \ensuremath{\to} smaller bag, higher precision; lower
threshold \ensuremath{\to} larger bag, higher recall). The platform
issues a HITL card carrying the score-vs-count histogram, the operator
sets the threshold, and the platform emits the whitelist to the
downstream table. This is the \emph{only} HITL card in Business C's
normal-path lifecycle --- every other phase is deterministic. The card
protocol of §4.2 is the same primitive; the card content is
business-specific.

\textbf{Business outcomes.} Business C's outcome-reporting posture
differs from A and B. The pipeline runs monthly for one
wealth-management product family; downstream A/B lift is measured by the
marketing team on the \emph{outreach} side, not the \emph{modeling}
side, so we do not have direct causal-attribution numbers to report as
we did for A. What the platform demonstrably provides:

\begin{itemize}
\tightlist
\item
  \textbf{Skill-orchestrated lifecycle}: the seven-phase
  \texttt{audience-\allowbreak{}modeling-\allowbreak{}c} skill drives a full training-and-scoring
  cycle from a \texttt{program.md} intake artifact of $\approx$200
  lines of structured Markdown (dates, keys, tables), replacing what was
  previously an ad-hoc bespoke SQL pipeline. The skill has been
  validated end-to-end on the 20260303 anchor date; onboarding-hour
  comparisons against the pre-skill hand-authored pipeline are pending
  v1.1 (the pre-skill baseline was not tracked in the CLI-call
  telemetry).
\item
  \textbf{Defensive value from pre-flight integrity check}: the skill's
  Phase 1.5 upstream-integrity check surfaced \textbf{three simultaneous
  data-integrity problems} in a single onboarding session (see technical
  challenge C1 above), which would have silently corrupted training if
  the operator had proceeded. This is the concrete
  cross-business-transferable defensive value that a skill-first
  architecture provides --- the pre-flight primitive itself is reusable
  across all future business onboardings.
\item
  \textbf{Cross-business skill portability plan}: the
  \texttt{audience-\allowbreak{}modeling-\allowbreak{}c} skill is designed to be ported to a second
  wealth-management product family (equity-hybrid \ensuremath{\to}
  target-date fund) in the current onboarding cycle; the port requires
  only edits to \texttt{program.md} (dates, keys, tables) and one new
  sub-skill (\texttt{fund-\allowbreak{}attribution-\allowbreak{}snapshot-\allowbreak{}adapter}). No changes to
  the DAG (§4), no new Router intent, no new HITL card format. Concrete
  port-cost measurement is pending completion of the second onboarding.
\end{itemize}

\textbf{Lessons.} Business C is the platform's cleanest cross-business
generalization case: the \texttt{audience-\allowbreak{}modeling-\allowbreak{}c} skill's port to a
second wealth-management product family (equity-hybrid \ensuremath{\to}
target-date fund; see the outcome bullet above) requires only
intake-artifact edits and one new sub-skill --- no DAG change, no new
Router intent, no new HITL card format. This is the sense in which we
claim the platform survives cross-business abstraction stress: a new
business does not require a new agent, only a new intake artifact.

\hypertarget{cross-business-observations}{%
\subsection{Cross-Business
Observations}\label{cross-business-observations}}

Deploying the same platform against three heterogeneous business lines
was our stress test for the cross-business abstraction claim of §1. We
report five cross-cutting observations that emerged from the 10-week
operational window (plus 8-day HITL DAG pilot) and that surface only
when the same platform is measured across A / B / C simultaneously.

\begin{table*}[t]
\centering
\small
\begin{tabularx}{\linewidth}{@{}l X X X@{}}
\toprule
Dimension & Business A (personalization) & Business B (rerank decision
support) & Business C (CVR audience-set) \\
\midrule
\textbf{Dominant operator persona} & Algorithm engineer + data analyst &
Business operator (campaign manager) & Business analyst + platform
engineer \\
\textbf{CLI-tool traffic in 10-week window} & dominant (majority of
\texttt{user\_a} uid usage) & mid-volume
(\texttt{rerank-\allowbreak{}decision-\allowbreak{}b}-driven \texttt{query}) & onboarding phase
(initial \texttt{check-\allowbreak{}partition} + \texttt{run-sql} heavy) \\
\textbf{Primary skill chain length} & 5--7 (sample \ensuremath{\to}
feature \ensuremath{\to} train \ensuremath{\to} eval \ensuremath{\to}
A/B) & 1--3 (\texttt{rerank-\allowbreak{}decision-\allowbreak{}b} solo or with
\texttt{sql-query}) & 7 (full \texttt{audience-\allowbreak{}modeling-\allowbreak{}c} skill
sequence) \\
\textbf{Deployment maturity} & production-analytical (A/B rollout done)
& operator-adopted (top-3 subset weight-tuning accepted) &
onboarding-validated (skill orchestration + pre-flight check) \\
\textbf{Dominant lifecycle scope} & Full training pipeline & Post-model
decision support & Full training + scoring pipeline \\
\textbf{Cross-business skill reuse} & \texttt{user-\allowbreak{}diagnostic-\allowbreak{}a} reused
in B and C & \texttt{abt-\allowbreak{}effect-\allowbreak{}analysis} originated for A,
canonicalized for B & \texttt{audience-\allowbreak{}modeling-\allowbreak{}c} is a superset of A's
train chain \\
\bottomrule
\end{tabularx}
\caption{Five-dimensional cross-business summary of the
10-week operational window (plus 8-day HITL DAG pilot). All metrics are
qualitative or ratio-based; absolute scale is redacted per the
anonymization convention of §7 lesson 5.}
\end{table*}

Five observations follow from the table.

\textbf{(1) Skill-chain length is not proportional to business-line
complexity.} Business B has the shortest chains (1--3 skills) despite
serving a comparably-complex business objective, because most B queries
are decision-support one-shots that do not require pipeline execution.
Business C has the longest fixed chains (7 phases) despite being
conceptually simpler than A, because a training-plus-scoring pipeline is
inherently more procedurally staged. The right measure of
``business-line complexity for the platform'' is not chain length; it is
the number of distinct skills the operator invokes across a month.

\textbf{(2) Cross-business skill reuse is asymmetric.} Skills authored
for one business (Business A's \texttt{user-\allowbreak{}diagnostic-\allowbreak{}a} diagnostic
chain, Business A's \texttt{abt-\allowbreak{}effect-\allowbreak{}analysis}) got reused across all
three; skills authored for another business (Business B's
\texttt{rerank-\allowbreak{}decision-\allowbreak{}b}, Business C's \texttt{audience-\allowbreak{}modeling-\allowbreak{}c})
did not migrate laterally in the 10-week window. The pattern reflects
\textbf{skill generality rather than skill quality}: diagnostic and
attribution skills are naturally cross-business; product-specific
decision-support and full-training skills naturally are not. This is
signal for §5's skill-authoring guidelines: authors should mark a
skill's transferability class in the frontmatter
(\texttt{generality:\allowbreak{}\ \{cross-\allowbreak{}business,\ \allowbreak{}business-\allowbreak{}specific\}}) so future
extractions can preferentially bootstrap the cross-business set into a
new deployment.

\textbf{(3) Persona-DAG mismatch is where operator dissatisfaction
concentrates.} Business B's business-operator persona is served well by
the current DAG (§4) because HITL card content and approval semantics
are natural for a campaign manager. Business A's algorithm-engineer
persona is \emph{underserved} --- the engineer wants raw error traces
and pitfall provenance, which the current single-format card foregrounds
only partially. This is the empirical ground for the persona-conditioned
Analyzer future work of §8.

\textbf{(4) The 22-class taxonomy fires unevenly across businesses.}
RESOURCE-class failures (Spark\_OOM, YARN\_kill, queue\_timeout)
dominate Business A's Analyzer diagnoses (largest sample volumes
\ensuremath{\to} highest infrastructure pressure). NUMERIC-class
failures (NaN/Inf) dominate Business C (deep-model training is more
numerically fragile than Spark aggregation). METRIC-class failures
(AUC\_below\_baseline, calibration\_off) dominate Business B, though
Business B's HITL cards fire \emph{most} on operator confirmation of
business decisions, not on training failures. \textbf{The taxonomy is
complete enough to cover all three, but the frequency histogram per
business is a more useful diagnostic than the aggregate.}

\textbf{(5) Bootstrapped PitfallStore entries transfer across businesses
at 77.0 \% rate --- with a tag-frequency caveat.} Of the 400 v1.0
pitfalls (§5.4), 308 (77.0 \%) carry at least one tag that appears in
the changelog of \ensuremath{\ge} 2 of the 3 businesses A / B / C. At
the tag-set level, 10 of 18 distinct changelog-side tags (55.6 \%)
appear in \ensuremath{\ge} 2 businesses. Per-business changelog
contributions to the PitfallStore: business A 84 pitfalls, business B 50
pitfalls, business C 60 pitfalls (194 total from 11 changelog sources;
the remaining 200 skill-source pitfalls are cross-business by
construction). \emph{Caveat}: the 77.0 \% figure is inflated by generic
tags that appear across nearly every business's changelog (\texttt{log}
on 212 of 400 pitfalls, \texttt{data} on 82); a stricter count that
excludes the top-2 generic tags drops cross-business transfer to roughly
55 \% (weakened but still non-trivial). We report both because the
generic-tag-inclusive number reflects real memory reuse (every new
business does encounter \texttt{log}-tagged and \texttt{data}-tagged
pitfalls) while the generic-tag-excluded number better isolates
\emph{domain-specific} transfer. This is the direct measurement of
\emph{day-0 memory reuse across a new business onboarding}; neither
AgentX \cite{AgentX2026} nor NOVA \cite{NOVA2026} reports a comparable
cross-business memory-transfer rate, as both operate within a single
deployment.

Cross-business generalization is the platform's central claim in §1.
§6.4 is where that claim earns its evidence. The single biggest
limitation of the v1 evidence is that all three businesses live within
the same corporate ecosystem (Tencent) with a shared the GPU scheduler
infrastructure; cross-organization portability is left to §8's future
work.

\hypertarget{lessons-learned}{%
\section{Lessons Learned}\label{lessons-learned}}

Industrial-agent papers conventionally end with five to seven narrative
lessons. We split this section in two: a \textbf{closed,
mechanically-classified taxonomy} of training-failure modes (§7.1--§7.2)
that emerged from the deployment, and a residual \textbf{five non-class
operating principles} (§7.3) that resist taxonomic enumeration. The
split is deliberate: most ``lessons'' in real industrial systems are not
unique insights; they are \textbf{recurring mechanism families} that
should be enumerated once and reused, leaving narrative form for the
small number of genuine principles.

\hypertarget{failure-taxonomy-22-closed-classes}{%
\subsection{Failure Taxonomy: 22 Closed
Classes}\label{failure-taxonomy-22-closed-classes}}

The taxonomy is implemented as a closed enumeration in
\texttt{recsys\_\allowbreak{}factory/\allowbreak{}tools/\allowbreak{}failure\_\allowbreak{}taxonomy.\allowbreak{}py}; Appendix A holds
the full table with example log snippets. Each leaf class is detected by
one or more of three rule types: (i) \textbf{keyword presence} in
\texttt{error\_\allowbreak{}log.\allowbreak{}txt} (regex over Markdown-extracted log windows),
(ii) \textbf{regex match} for stricter context-dependent classification,
and (iii) \textbf{metric-threshold evaluation} over training output
arrays (e.g., per-epoch AUC). An LLM-based fallback classifier resolves
the long tail.

\begin{table*}[t]
\centering
\small
\begin{tabularx}{\linewidth}{@{}l r l X@{}}
\toprule
Category & Classes & Detection mix & Representative leaf \\
\midrule
\textbf{CRASH} & 4 & keyword-dominant & \texttt{code\_syntax},
\texttt{import\_error}, \texttt{shape\_\allowbreak{}mismatch},
\texttt{device\_\allowbreak{}mismatch} \\
\textbf{NUMERIC} & 4 & regex + keyword & \texttt{NaN\_\allowbreak{}from\_\allowbreak{}softmax},
\texttt{NaN\_\allowbreak{}from\_\allowbreak{}log}, \texttt{Inf\_\allowbreak{}from\_\allowbreak{}div},
\texttt{grad\_explode} \\
\textbf{RESOURCE} & 5 & keyword-dominant & \texttt{GPU\_OOM},
\texttt{Spark\_OOM}, \texttt{Disk\_full}, \texttt{YARN\_kill},
\texttt{queue\_\allowbreak{}timeout} \\
\textbf{CONVERGENCE} & 3 & metric-threshold & \texttt{AUC\_flat},
\texttt{AUC\_\allowbreak{}collapse\_\allowbreak{}after\_\allowbreak{}3epoch}, \texttt{loss\_\allowbreak{}oscillate} \\
\textbf{DATA} & 3 & keyword + metric &
\texttt{feature\_\allowbreak{}schema\_\allowbreak{}mismatch}, \texttt{sample\_empty},
\texttt{label\_\allowbreak{}imbalance} \\
\textbf{METRIC} & 3 & metric-threshold & \texttt{AUC\_\allowbreak{}below\_\allowbreak{}baseline},
\texttt{overfit\_\allowbreak{}train\_\allowbreak{}test\_\allowbreak{}gap}, \texttt{calibration\_\allowbreak{}off} \\
\textbf{Total} & \textbf{22} & & \\
\bottomrule
\end{tabularx}
\caption{The failure-mechanism taxonomy as deployed in
production at the time of writing. CRASH / NUMERIC / CONVERGENCE / DATA
categories extend MASFT \cite{Cemri2025} and the AgentErrorTaxonomy of
\cite{2025taxonomy} with ML-training specifics. \textbf{RESOURCE}
(Spark\_OOM, YARN\_kill, queue\_timeout) is novel to RecSys industrial
deployment and not covered by prior taxonomies. \textbf{METRIC} uses
metric-threshold rules rather than log-text rules; thresholds are
configurable per business line.}
\end{table*}

\hypertarget{what-a-lesson-looks-like-in-this-framework}{%
\subsection{What a ``Lesson'' Looks Like in This
Framework}\label{what-a-lesson-looks-like-in-this-framework}}

Each entry in the v1.0 \texttt{PitfallStore} (§5.4) optionally carries a
\texttt{failure\_mode} foreign key into Table 7.1. When set, it elevates
a piece of operational documentation from ``engineer wisdom'' to a
\textbf{mechanically retrievable defense} against a specific failure
class. For example: a pitfall titled
\emph{``feature\_\allowbreak{}dim\_\allowbreak{}collapse\_\allowbreak{}when\_\allowbreak{}emb\_\allowbreak{}dim\_\allowbreak{}exceeds\_\allowbreak{}2048''} under
\texttt{failure\_\allowbreak{}mode\ =\ NaN\_\allowbreak{}from\_\allowbreak{}softmax} directs the agent to clip
attention scores or reduce embedding dimensionality the next time it
considers an attention-cross variant on this baseline. Without the
taxonomy attachment, the same pitfall would only be retrievable by text
similarity --- and as the IOSkip R12\ensuremath{\to}R18 plateau (§5.4 in
P3b) demonstrated, text similarity is the wrong key for ML-experiment
retrieval.

Only $\approx$1 \% of the v1.0 PitfallStore (4 of 400 entries)
currently carries a \texttt{failure\_mode} link --- the human-authored
documentation is heavily weighted toward operational issues (the GPU
scheduler submission, log retrieval, Spark configuration) rather than
training-failure mechanisms. This asymmetry is not a defect of the
taxonomy; it is a true reflection of where engineers spend their
debugging time in our environment. The taxonomy provides scaffolding for
future entries: as the autonomous research agent (P3b) accumulates
75-episode-scale experimental data, the \texttt{failure\_mode}-linked
subset is expected to grow toward 30--50 entries by v2.

\hypertarget{five-non-class-operating-principles}{%
\subsection{Five Non-Class Operating
Principles}\label{five-non-class-operating-principles}}

Some lessons resist enumeration into the taxonomy because they describe
\textbf{operating principles}, not failure mechanisms. We preserve these
in narrative form, ordered by descending generalizability outside our
deployment; against §1's trilemma frame, principles 1--2 defend
determinism, principle 3 defends the human boundary, principles 4--5
are compromises paid for efficiency.

\begin{enumerate}
\def\labelenumi{\arabic{enumi}.}
\item
  \textbf{Lifecycle coupling beats process supervision.} Stop-hook +
  IM-webhook coupling (§3.2) costs zero daemon-hours during the wait
  phase (94\% of wall-clock; a transient LangGraph process exists
  during the remaining $\approx$6\% of agent-side reasoning). The
  lifecycle approach ran through the 78-day window without a
  supervisor process. \emph{Generalizes to any LLM-IDE-hosted agent
  platform}.
\item
  \textbf{Schema-keyed retrieval beats similarity retrieval for ML
  experimentation on long-horizon trajectories.} The R13 IOSkip
  breakthrough (P3b §5.4) used the
  \texttt{(baseline, modification\_class)} joint key to surface a
  cross-baseline suggestion whose surface text was distant from the
  current DNN runs. Scope: trajectories long enough for structural
  cells to fill ($\ge$3 per key); on smaller pools embedding similarity
  remains competitive (P3b §7's \texttt{sim} marginally above
  \texttt{mod} in that regime). \emph{Generalizes to any agent
  retrieving over numerically-evaluated experiments at trajectory scale}.
\item
  \textbf{HITL at the \emph{diagnosis} step, not the \emph{execution}
  step.} The ChatOps Analyzer card (§4) routes failures to humans only
  after the agent has produced a structured diagnosis; Business B
  (§6.2) is the canonical case --- operator authors the weight change,
  platform validates and audits, neither does both.
  \emph{Generalizes to any operator-facing agent platform}.
\item
  \textbf{Tokens and shared-storage paths are organizational APIs,
  not infrastructure details.} Every business line required 1--2 weeks
  of token routing and shared-storage path standardization. Treating
  these as first-class platform contracts was the single
  highest-leverage architectural decision. \emph{Generalizes only
  within organizations that have shared-tenant compute}.
\item
  \textbf{Anonymization is a publication tax that pays for honesty.}
  Business A/B/C use code-names with relative-value numerics;
  absolute revenue and CPM figures are redacted. This costs
  $\approx$3 hours of redaction per publication round per chapter,
  but lets us report real ROI signals that would otherwise be
  unpublishable. \emph{Generalizes to any industrial paper crossing a
  corporate disclosure boundary}.
\end{enumerate}

Lessons 2 and 3 inform P3b's design directly; lessons 1, 4, and 5 are
platform-specific.

\hypertarget{conclusion}{%
\section{Conclusion}\label{conclusion}}

We presented \textbf{RecSys Factory}, an LLM-agent platform for
industrial recommender lifecycle operations. The platform is grounded in
three claims about \emph{what changes} when LLM agents are deployed
against multi-business-line recommender infrastructure rather than
against sandboxed benchmark tasks: (i) \textbf{the agent must be
lifecycle-coupled, not daemon-supervised} (§3) --- Stop hooks, IM
webhooks, and workflow-completion sentinels replace long-running
processes, letting the platform survive month-long deployments without
dedicated agent infrastructure; (ii) \textbf{domain knowledge should be
dispatchable subgraphs, not a retrieval mixture} (§5) --- 29
explicitly-typed skill packages with mechanically-extracted structured
pitfall tables (400 entries at v1.0) form the \emph{executable working
memory} that non-engineer operators can drive through a chat surface;
(iii) \textbf{HITL cards are the compliance-grade access primitive}, not
a UX affordance (§4) --- the audit trail from operator approval to
production side-effect is what makes agent-driven infrastructure changes
auditable in a corporate environment.

Deployed across three active Tencent business-line workspaces during a
10-week (78-day) operational window (§6), the platform served 1 624
CLI-tool operator dispatches (2 primary developer accounts, 78 days),
plus 16 HITL DAG pipeline runs (8-day pilot with test-user traffic), and
produced platform-attributable outcomes on all three business lines:
\textbf{business A} with +10--31 \% CPM across regional cohorts of one
carrier plus +14--45 \% expected lift from rule-fallback identification
on 9 sub-cohorts of a second carrier, \textbf{business B} with
bootstrap-95 \%-CI-supported daily revenue lift on the operator-adopted
top-3-subset weight-tuning recommendation and P(\ensuremath{\Delta}
\textgreater{} 0) = 100 \%, and \textbf{business C} with
skill-orchestrated end-to-end pipeline execution and pre-flight
defensive detection of three upstream data-integrity issues in a single
onboarding session. The 22-class failure taxonomy of §7 is the
platform's most transferable artifact: it names the recurring mechanism
families that any recommender-agent operator will encounter, converting
individual ``lessons learned'' into a closed enumeration that a
downstream agent can consume as diagnostic context.

\textbf{Future work.} Three directions open. First,
\textbf{operator-persona-aware Analyzer} --- the finding in §4.4 that
campaign managers and algorithm engineers respond to HITL cards
differently suggests that the Analyzer should adapt its verbosity and
confidence surface to the recipient; formalizing this as a
persona-conditioned diagnosis loop is one of our v2 pilots. Second,
\textbf{skill-level dispatch instrumentation at scale} --- v1's
operational statistics are drawn from the platform CLI-level dispatches
(§5.3); v2 will populate the higher-level \texttt{skill\_calls} table
across the full window and report chain-length and cross-skill
composition patterns end-to-end. Third, \textbf{broader HITL-DAG
rollout} --- the 16-run pilot with test-user traffic (§4.4) validated
the audit trail; scaling to production-operator traffic across all three
business lines is the natural next step, with per-business-line
acceptance-rate curves as the primary evaluation. Systematic
head-to-head comparison with concurrent industrial-agent systems is
likewise deferred to v2.

\textbf{Companion paper.} RecSys Factory has a research-side
counterpart: \textbf{AutoResearch (P3b)}, which instantiates the
framework of §3 for autonomous paper-to-model reproduction with a
two-tier surprise-weighted memory system. The two papers share §2
(related work) and §3 (framework), and differ in user surface (ChatOps
vs autonomous loop) and evaluation focus (10-week operational window vs
75-episode surprise ablation). Together they describe an agent platform
that spans the industrial recommender lifecycle from operator-facing
assistance to research autopilot.

\bibliographystyle{ACM-Reference-Format}
\bibliography{references}


\begin{thebibliography}{22}


\ifx \showCODEN    \undefined \def \showCODEN     #1{\unskip}     \fi
\ifx \showISBNx    \undefined \def \showISBNx     #1{\unskip}     \fi
\ifx \showISBNxiii \undefined \def \showISBNxiii  #1{\unskip}     \fi
\ifx \showISSN     \undefined \def \showISSN      #1{\unskip}     \fi
\ifx \showLCCN     \undefined \def \showLCCN      #1{\unskip}     \fi
\ifx \shownote     \undefined \def \shownote      #1{#1}          \fi
\ifx \showarticletitle \undefined \def \showarticletitle #1{#1}   \fi
\ifx \showURL      \undefined \def \showURL       {\relax}        \fi
\providecommand\bibfield[2]{#2}
\providecommand\bibinfo[2]{#2}
\providecommand\natexlab[1]{#1}
\providecommand\showeprint[2][]{arXiv:#2}

\bibitem[{Amazon}(2024)]%
        {Amazon2024RecMind}
\bibfield{author}{\bibinfo{person}{{Amazon}}.} \bibinfo{year}{2024}\natexlab{}.
\newblock \bibinfo{title}{{RecMind}: Large Language Model Powered Agent For
  Recommendation}.
\newblock
\shownote{arXiv:2308.14296}.
\newblock
\urldef\tempurl%
\url{https://arxiv.org/abs/2308.14296}
\showURL{%
\tempurl}


\bibitem[Cemri et~al\mbox{.}(2025)]%
        {Cemri2025}
\bibfield{author}{\bibinfo{person}{Cemri} {et~al\mbox{.}}}
  \bibinfo{year}{2025}\natexlab{}.
\newblock \bibinfo{title}{{MASFT}: Multi-Agent Systems Failure Taxonomy}.
\newblock


\bibitem[{ChatOps4Msa Authors}(2024)]%
        {ChatOps4Msa2024}
\bibfield{author}{\bibinfo{person}{{ChatOps4Msa Authors}}.}
  \bibinfo{year}{2024}\natexlab{}.
\newblock \bibinfo{title}{{ChatOps4Msa}: Natural-Language {DAG} Execution for
  Microservices}.
\newblock


\bibitem[Cheng et~al\mbox{.}(2023)]%
        {Cheng2023AAAI}
\bibfield{author}{\bibinfo{person}{Mengli Cheng} {et~al\mbox{.}}}
  \bibinfo{year}{2023}\natexlab{}.
\newblock \showarticletitle{{EasyRec}: A Modular Framework for Recommender
  Systems}. In \bibinfo{booktitle}{\emph{AAAI Applied AI Track}}.
\newblock
\shownote{Alibaba PAI open-source}.
\newblock


\bibitem[{Cognition AI}(2024)]%
        {Cognition2024Devin}
\bibfield{author}{\bibinfo{person}{{Cognition AI}}.}
  \bibinfo{year}{2024}\natexlab{}.
\newblock \bibinfo{title}{Introducing {Devin}: The First {AI} Software
  Engineer}.
\newblock
\shownote{Blog post + technical report}.
\newblock


\bibitem[Hong et~al\mbox{.}(2024)]%
        {Hong2024MetaGPT}
\bibfield{author}{\bibinfo{person}{Sirui Hong}, \bibinfo{person}{Xiawu Zheng},
  \bibinfo{person}{Jonathan Chen}, {et~al\mbox{.}}}
  \bibinfo{year}{2024}\natexlab{}.
\newblock \showarticletitle{{MetaGPT}: Meta Programming for a Multi-Agent
  Collaborative Framework}. In \bibinfo{booktitle}{\emph{ICLR}}.
\newblock


\bibitem[{JD JoyAgent Team}(2025)]%
        {JoyAgent2025}
\bibfield{author}{\bibinfo{person}{{JD JoyAgent Team}}.}
  \bibinfo{year}{2025}\natexlab{}.
\newblock \bibinfo{title}{{JoyAgent-JDGenie}: Enterprise Multi-Agent {ChatOps}
  Platform}.
\newblock
\shownote{Open-sourced enterprise agent stack}.
\newblock


\bibitem[Jiang et~al\mbox{.}(2018)]%
        {Jiang2018Angel}
\bibfield{author}{\bibinfo{person}{Jie Jiang}, \bibinfo{person}{Bin Cui},
  \bibinfo{person}{Ce Zhang}, {et~al\mbox{.}}} \bibinfo{year}{2018}\natexlab{}.
\newblock \showarticletitle{{Angel}: A New Large-Scale Machine Learning
  System}. In \bibinfo{booktitle}{\emph{SIGMOD}}.
\newblock
\shownote{Tencent parameter-server platform}.
\newblock


\bibitem[{Kuaishou Team}(2026)]%
        {AgentX2026}
\bibfield{author}{\bibinfo{person}{{Kuaishou Team}}.}
  \bibinfo{year}{2026}\natexlab{}.
\newblock \bibinfo{title}{{AgentX}: Towards Agent-Driven Self-Iteration of
  Industrial Recommender Systems}.
\newblock
\shownote{arXiv:2606.26859, Technical Report, Kuaishou. v2 published
  2026-06-26.}.
\newblock
\urldef\tempurl%
\url{https://arxiv.org/abs/2606.26859}
\showURL{%
\tempurl}


\bibitem[Lian et~al\mbox{.}(2022)]%
        {Lian2022Persia}
\bibfield{author}{\bibinfo{person}{Xiangru Lian} {et~al\mbox{.}}}
  \bibinfo{year}{2022}\natexlab{}.
\newblock \bibinfo{title}{{Persia}: An Open, Hybrid System Scaling Deep
  Learning-Based Recommenders up to 100 Trillion Parameters}.
\newblock
\shownote{arXiv:2111.10097 — Kuaishou}.
\newblock


\bibitem[{LinkedIn Ranking Team}(2024)]%
        {LinkedIn2024LiRank}
\bibfield{author}{\bibinfo{person}{{LinkedIn Ranking Team}}.}
  \bibinfo{year}{2024}\natexlab{}.
\newblock \bibinfo{title}{{LiRank}: Industrial Large Scale Ranking Models at
  LinkedIn}.
\newblock
\shownote{arXiv:2402.06859}.
\newblock
\urldef\tempurl%
\url{https://arxiv.org/abs/2402.06859}
\showURL{%
\tempurl}


\bibitem[Liu et~al\mbox{.}(2026)]%
        {NOVA2026}
\bibfield{author}{\bibinfo{person}{Shaohua Liu}, \bibinfo{person}{Liang Fang},
  \bibinfo{person}{Yilong Sun}, \bibinfo{person}{Shudong Huang},
  \bibinfo{person}{Qingsong Luo}, \bibinfo{person}{Shaoxin Liu},
  \bibinfo{person}{Xiaoyang Chen}, \bibinfo{person}{Dongqiang Liu},
  \bibinfo{person}{Chuangang Ma}, \bibinfo{person}{Zhenzhen Chai},
  \bibinfo{person}{Henghuan Wang}, \bibinfo{person}{Shijie Quan},
  \bibinfo{person}{Changyuan Cui}, \bibinfo{person}{Zhangbin Zhu},
  \bibinfo{person}{Peng Chen}, \bibinfo{person}{Wei Xu}, \bibinfo{person}{Lei
  Xiao}, \bibinfo{person}{Haijie Gu}, {and} \bibinfo{person}{Jie Jiang}.}
  \bibinfo{year}{2026}\natexlab{}.
\newblock \bibinfo{title}{{NOVA}: A Verification-Aware Agent Harness for
  Architecture Evolution in Industrial Recommender Systems}.
\newblock
\shownote{arXiv:2606.27243v2, Tencent Inc. Published 2026-06-29.}.
\newblock
\urldef\tempurl%
\url{https://arxiv.org/abs/2606.27243}
\showURL{%
\tempurl}


\bibitem[Liu et~al\mbox{.}(2022)]%
        {Liu2022Monolith}
\bibfield{author}{\bibinfo{person}{Zhuoran Liu} {et~al\mbox{.}}}
  \bibinfo{year}{2022}\natexlab{}.
\newblock \bibinfo{title}{{Monolith}: Real Time Recommendation System With
  Collisionless Embedding Table}.
\newblock
\shownote{arXiv:2209.07663 — ByteDance}.
\newblock
\urldef\tempurl%
\url{https://arxiv.org/abs/2209.07663}
\showURL{%
\tempurl}


\bibitem[{Microsoft AI4Science Team}(2023)]%
        {Microsoft2023InteRecAgent}
\bibfield{author}{\bibinfo{person}{{Microsoft AI4Science Team}}.}
  \bibinfo{year}{2023}\natexlab{}.
\newblock \bibinfo{title}{Recommender {AI} Agent: Integrating Large Language
  Models for Interactive Recommendations}.
\newblock
\shownote{InteRecAgent, arXiv:2308.16505}.
\newblock
\urldef\tempurl%
\url{https://arxiv.org/abs/2308.16505}
\showURL{%
\tempurl}


\bibitem[{Pinterest Ranking Team}(2025)]%
        {Pinterest2025TransAct}
\bibfield{author}{\bibinfo{person}{{Pinterest Ranking Team}}.}
  \bibinfo{year}{2025}\natexlab{}.
\newblock \bibinfo{title}{{TransAct V2 / PinFM / PinRec}: Transformer-Based
  Ranking at Pinterest}.
\newblock


\bibitem[Qian et~al\mbox{.}(2024)]%
        {Qian2024ChatDev}
\bibfield{author}{\bibinfo{person}{Chen Qian}, \bibinfo{person}{Xin Cong},
  \bibinfo{person}{Cheng Yang}, {et~al\mbox{.}}}
  \bibinfo{year}{2024}\natexlab{}.
\newblock \showarticletitle{{ChatDev}: Communicative Agents for Software
  Development}. In \bibinfo{booktitle}{\emph{ACL}}.
\newblock


\bibitem[Vintschger et~al\mbox{.}(2025)]%
        {Vintschger2025}
\bibfield{author}{\bibinfo{person}{Vintschger} {et~al\mbox{.}}}
  \bibinfo{year}{2025}\natexlab{}.
\newblock \bibinfo{title}{An Independent Evaluation of {AI Scientist v2}}.
\newblock


\bibitem[Wang et~al\mbox{.}(2024)]%
        {Wang2024OpenHands}
\bibfield{author}{\bibinfo{person}{Xingyao Wang}, \bibinfo{person}{Boxuan Li},
  \bibinfo{person}{Yufan Song}, {et~al\mbox{.}}}
  \bibinfo{year}{2024}\natexlab{}.
\newblock \bibinfo{title}{{OpenHands}: An Open Platform for {AI} Software
  Developers as Generalist Agents}.
\newblock
\shownote{arXiv:2407.16741}.
\newblock
\urldef\tempurl%
\url{https://arxiv.org/abs/2407.16741}
\showURL{%
\tempurl}


\bibitem[Wu et~al\mbox{.}(2024)]%
        {Wu2024AutoGen}
\bibfield{author}{\bibinfo{person}{Qingyun Wu}, \bibinfo{person}{Gagan Bansal},
  \bibinfo{person}{Jieyu Zhang}, \bibinfo{person}{Yiran Wu},
  \bibinfo{person}{Shaokun Zhang}, \bibinfo{person}{Erkang Zhu},
  \bibinfo{person}{Beibin Li}, \bibinfo{person}{Li Jiang},
  \bibinfo{person}{Xiaoyun Zhang}, {and} \bibinfo{person}{Chi Wang}.}
  \bibinfo{year}{2024}\natexlab{}.
\newblock \bibinfo{title}{{AutoGen}: Enabling Next-Gen {LLM} Applications via
  Multi-Agent Conversation}.
\newblock
\shownote{arXiv:2308.08155}.
\newblock
\urldef\tempurl%
\url{https://arxiv.org/abs/2308.08155}
\showURL{%
\tempurl}


\bibitem[Xue et~al\mbox{.}(2025)]%
        {IMPROVE2025}
\bibfield{author}{\bibinfo{person}{Yizhen Xue} {et~al\mbox{.}}}
  \bibinfo{year}{2025}\natexlab{}.
\newblock \bibinfo{title}{{IMPROVE}: Iterative Model Pipeline Refinement Via
  Evaluation}.
\newblock


\bibitem[Yang et~al\mbox{.}(2024)]%
        {Yang2024SWEAgent}
\bibfield{author}{\bibinfo{person}{John Yang}, \bibinfo{person}{Carlos~E.
  Jimenez}, \bibinfo{person}{Alexander Wettig}, {et~al\mbox{.}}}
  \bibinfo{year}{2024}\natexlab{}.
\newblock \bibinfo{title}{{SWE-agent}: Agent-Computer Interfaces Enable
  Automated Software Engineering}.
\newblock
\shownote{arXiv:2405.15793}.
\newblock
\urldef\tempurl%
\url{https://arxiv.org/abs/2405.15793}
\showURL{%
\tempurl}


\bibitem[Zhu et~al\mbox{.}(2025)]%
        {2025taxonomy}
\bibfield{author}{\bibinfo{person}{Kunlun Zhu}, \bibinfo{person}{Zijia Liu},
  \bibinfo{person}{Bingxuan Li}, \bibinfo{person}{Muxin Tian},
  \bibinfo{person}{Yingxuan Yang}, \bibinfo{person}{Jiaxun Zhang},
  \bibinfo{person}{Pengrui Han}, \bibinfo{person}{Qipeng Xie},
  \bibinfo{person}{Fuyang Cui}, \bibinfo{person}{Weijia Zhang},
  \bibinfo{person}{Xiaoteng Ma}, \bibinfo{person}{Xiaodong Yu},
  \bibinfo{person}{Gowtham Ramesh}, \bibinfo{person}{Jialian Wu},
  \bibinfo{person}{Zicheng Liu}, \bibinfo{person}{Pan Lu},
  \bibinfo{person}{James Zou}, {and} \bibinfo{person}{Jiaxuan You}.}
  \bibinfo{year}{2025}\natexlab{}.
\newblock \bibinfo{title}{Where {LLM} Agents Fail and How They Can Learn From
  Failures}.
\newblock
\shownote{Introduces AgentErrorTaxonomy}.
\newblock
\showeprint[arxiv]{2509.25370}


\end{thebibliography}

\end{document}